\documentclass[leqno,final,onefignum,onetabnum]{siamltexmm}
\usepackage{amsfonts}
\usepackage{amsmath}
\usepackage[dvips]{graphicx}
\usepackage{tabularx}
\usepackage{afterpage}
\usepackage[ruled]{algorithm2e}
\usepackage[colorlinks,breaklinks,pdfmark,dvips,bookmarks]{hyperref}
\usepackage{textcomp}

\def \imageDir {./Images/eps_images/}
\newcommand{\etal}{\emph{et al}.}
\newcommand{\cH}{\mathcal{H}}
\newcommand{\D}{\mathcal{D}}
\newcommand{\N}{\mathcal{N}}

\graphicspath{{./}}

\DeclareGraphicsExtensions{,.ps,.eps,.eps.gz}

\makeatletter
\makeatletter

\title{VIDEO INPAINTING OF COMPLEX SCENES} 

\author{
Alasdair Newson\footnotemark[2]\ \footnotemark[3]
\and Andr\'es Almansa \footnotemark[3]
\and Matthieu Fradet \footnotemark[2]
\and Yann Gousseau \footnotemark[3]
\and Patrick P\'erez \footnotemark[2]
}

\begin{document}
\maketitle
\newcommand{\slugmaster}{%
}

\renewcommand{\thefootnote}{\fnsymbol{footnote}}
\footnotetext[2]{Technicolor, 35570 Cesson-S\'evign\'e, France}
\footnotetext[3]{T\'el\'ecom ParisTech, CNRS LTCI, 75013 Paris, France}
\renewcommand{\thefootnote}{\arabic{footnote}}

\begin{abstract}
We propose an automatic video inpainting algorithm
which relies on the optimisation of a global, patch-based
functional. Our algorithm is able to deal with a variety
of challenging situations which naturally arise
in video inpainting, such as the correct reconstruction
of dynamic textures, multiple moving objects and
moving background. Furthermore, we achieve this in
an order of magnitude less execution time with respect
to the state-of-the-art. We are also able to achieve
good quality results on high definition videos.
Finally, we provide specific algorithmic
details to make implementation of our algorithm as easy as possible.
The resulting algorithm requires no segmentation or manual
input other than the definition of the inpainting mask,
and can deal with a wider variety of situations than
is handled by previous work.
\end{abstract}



\pagestyle{myheadings}
\thispagestyle{plain}
\markboth{NEWSON, ALMANSA, FRADET, GOUSSEAU AND P\'EREZ}{VIDEO INPAINTING}

\section{Introduction}
\label{sec:intro}
Advanced image and video editing techniques are increasingly common in the image processing and
computer vision world, and are also starting to be used in media entertainment.
One common and difficult task closely linked to the world of video editing
is image and video ``inpainting''. 
Generally speaking, this is the task of replacing
the content of an image or video with some other content which is visually pleasing.
This subject has been extensively studied in the case of images, to such an extent that
commercial image inpainting products destined for the general public are available, such as Photoshop's
``Content Aware fill'' \cite{PhotoshopCS5}. However,
while some impressive results have been obtained in the case of videos,
the subject has been studied far less extensively than image inpainting.
This relative lack of research can largely be attributed to high time complexity due to
the added temporal dimension. Indeed, it has only very recently become possible
to produce good quality inpainting results on high definition videos, and this only
in a semi-automatic manner. Nevertheless, high-quality video inpainting has many important and
useful applications such as film restoration, professional post-production in cinema and
video editing for personal use. For this reason, we believe that an automatic,
generic video inpainting algorithm would be extremely useful for both
academic and professional communities.

\subsection{Prior work}
\label{subsec:priorWork}
The generic goal of replacing areas of arbitrary shapes and sizes in
images by some other content was first presented by Masnou and Morel
in \cite{Masnou1998Level}. This method used level-lines to \emph{disocclude} the
region to inpaint. The term ``inpainting'' was first introduced
by Bertalmio \etal ~in \cite{Bertalmio2000Image}. Subsequently, a vast amount
of research was done in the area of image inpainting \cite{Bertalmio2011Inpainting}, and to a lesser
extent in \emph{video} inpainting.

Generally speaking, video inpainting algorithms belong to either
the ``object-based'' or ``patch-based'' category.
Object-based algorithms usually segment the video into moving foreground
objects and
background that is either still or displays simple motion.
These segmented image sequences are then inpainted using separate algorithms.
The background is often inpainted using image inpainting
methods such as \cite{Criminisi2003Object}, whereas
moving objects are often copied into the
occlusion as smoothly as possible.
Unfortunately, such methods include restrictive hypotheses
on the moving objects' motion, such as strict periodicity.
Some object-based methods include
\cite{Cheung2006Efficient,Jia2006Video,Ling2011Virtual}.

Patch-based methods are based on the intuitive idea of
copying and pasting small video ``patches'' (rectangular cuboids of video information)
into the occluded area. These patches are very useful as they
provide a practical way of encoding local texture, structure and motion (in the video case).

Patches were first introduced for texture synthesis in images \cite{Efros1999Texture},
and subsequently used with great success in image inpainting
\cite{Bornard2002Missing,Criminisi2003Object,Drori2003Fragmentbased}.
These methods copy and paste patches into the occlusion in a greedy
fashion, which means that no global coherence of the solution can be guaranteed.
This general approach was extended by Patwardhan \etal
~to the spatio-temporal case in \cite{Patwardhan2005}. In
\cite{Patwardhan2007Video}, this approach was further improved
so that moving cameras could be dealt with.
This is reflected by the fact that a good segmentation of the scene
into moving foreground objects and background is needed to produce good quality
results. This lack of
global coherence can be a drawback, especially for the correct inpainting
of moving objects.

Another method leading to a different family of
algorithms was presented by Demanet \etal ~in \cite{Demanet2003Image}.
The key insight here is that inpainting can be viewed as a labelling
problem: each occluded pixel can be associated with an unoccluded pixel,
and the final labels of the pixels result from a discrete optimisation process.
This idea was subsequently followed by a series of image inpainting methods
\cite{He2012Statistics,Komodakis2007Image,Liu2013ExemplarBased,Pritch2009Shiftmap} which use
optimisation techniques such as graph cuts algorithms \cite{Boykov2001Fast}, to
minimise a global patch-based functional.
The vector field representing the correspondences between occluded and
unoccluded pixels is referred to as the \emph{shift map} by Pritch \etal ~\cite{Pritch2009Shiftmap},
a term which we shall use in the current work.
This idea was extended to the video case
by Granados \etal ~in \cite{Granados2012How}. They
propose a semi-automatic algorithm which
optimises the spatio-temporal shift map.
This algorithm presents impressive results on higher resolution images than are previously
found in the literature (up to $1120 \times 754$ pixels).
However, in order to
reduce the large search space
and high time complexity of the optimisation method,
manual tracking of moving occluded
objects is required.
To the best of our knowledge, the inpainting results
of Granados \etal ~are the most advanced to date,
and we shall therefore compare our
algorithm with these results.

We also note the work of Herling and Broll \cite{Herling2012PixMix}
whose goal is ``diminished reality'', which considers the
inpainting task coupled with a tracking problem.
This is the only approach of which we are aware which inpaints videos in a real-time manner.
However, the method
relies on restrictive hypotheses on the nature of the
scene to inpaint and can therefore only deal with tasks such as removing a static object from a rigid platform.

Another family of patch-based video inpainting methods was introduced
in the seminal work of Wexler \etal ~\cite{Wexler2004Spacetime}.
This paper proposes an iterative method that may be seen as an heuristic to solve a
global optimisation problem.
This work is widely cited and
well-known in the video inpainting domain, mainly because it
ensures global coherency in an automatic manner.
This method is in fact closely linked to methods such as non-local
denoising \cite{Buades2005Nonlocal}. This link was also
noted in the work of Arias \etal ~\cite{Arias2011Variational},
which introduced a general non-local patch-based variational framework for inpainting
in the image case. In fact, the algorithm of Wexler \etal ~may be
seen as a special case of this framework.
We shall refer to this
general approach as the \emph{non-local patch-based} approach.
Darabi \etal ~have presented another variation on the work of Wexler \etal ~for image inpainting purposes in \cite{Darabi2012Image}.
In the video inpainting case, the high dimensionality of the problem makes such
approaches extremely slow, in particular due to the nearest neighbour search, requiring up to several days for a few seconds of VGA video.
This problem was considered in our previous work \cite{Newson2013Towards},
which represented a first step towards achieving high quality video inpainting results even on high resolution videos.
Given the flexibility and potential of the non-local patch-based approach,
we use it here to form the core of the proposed method. In order
to obtain an algorithm which can deal with a wide variety of complex situations,
we consider and propose solutions to some of the most important questions which arise in video inpainting.

\subsection{Outline and contributions of the paper}
\label{subsec:videoInpaintingClaims}

The ultimate goal of our work is to produce an automatic and generic
video inpainting algorithm which can deal with complex and varied
situations.
In Section~\ref{sec:frameworkNotation}, we present the variational framework
and the notations which we shall use in this paper. The proposed algorithm and
contributions are presented in Section~\ref{sec:method}.
These contributions can be summarised with the following points: 

\begin{itemize}
\item we greatly accelerate the nearest neighbour search,
using an extension of the PatchMatch algorithm \cite{Barnes2009PatchMatch} to the spatio-temporal case (Section~\ref{subsec:annSearch});
\item we introduce texture features to the patch distance in order to correctly inpaint video textures (Section~\ref{subsec:inpaintingTextures});
\item we deal with the problem of moving background (Section~\ref{subsec:movingBackgroundInpainting}) in videos,
using a robust affine estimation of the dominant motion in the video;
\item we describe our initialisation scheme (Section~\ref{subsec:inpaintingInitialisation}), which
is often left unspecified in previous work;
\item we give precise details concerning the implementation of the multi-resolution scheme (Section~\ref{subsec:multiResolutionDetails}).
\end{itemize}

As shown in Section \ref{sec:results}, the proposed algorithm produces high quality results in an automatic manner,
in a wide range of complex video inpainting situations: moving cameras, multiple
moving objects, changing background and dynamic video textures.
No pre-segmentation of the video is required. One of the most significant advantages of the proposed method
is that it deals with all these situations in a single framework, rather than
having to resort to separate algorithms for each case, as in \cite{Granados2012How}
and \cite{Granados2012Background} for instance.
In particular, the problem of reconstructing dynamic video textures
has not been previously addressed in other inpainting algorithms. This
is a worthwhile advantage since the problem of synthesising video
textures, which is usually done with dedicated algorithms, can be
achieved in this single, coherent framework. Finally, our algorithm
does not need any manual input other than the inpainting mask, and does not rely on foreground/background
segmentation, which is the case of many other approaches \cite{Cheung2006Efficient,Granados2012How,Jia2006Video,Ling2011Virtual,Patwardhan2007Video}.

We provide an implementation of our algorithm, which is available at the following address:
\url{http://www.telecom-paristech.fr/~gousseau/video_inpainting}.

\section{Variational framework and notation}
\label{sec:frameworkNotation}
\begin{figure}[!ht]
\centering
\includegraphics[width = 0.8\linewidth]{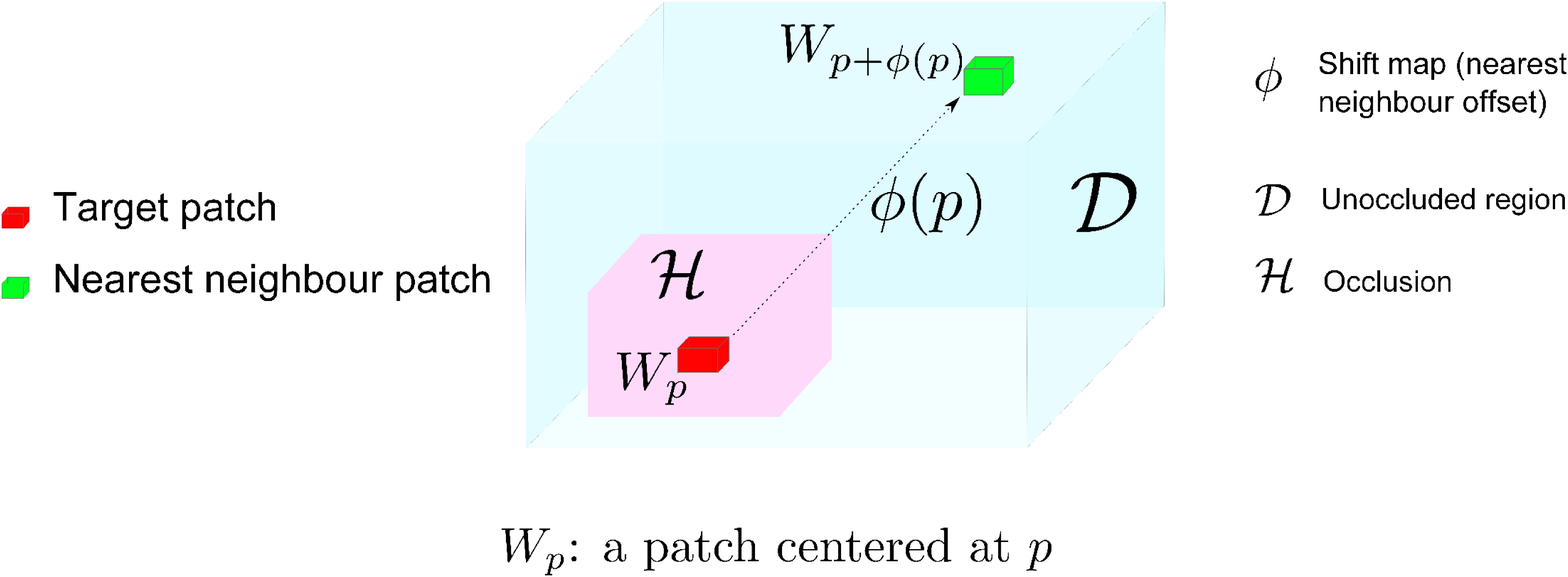}
\caption{Illustration of the notation used for proposed video inpainting algorithm.}
\label{fig:patchInpaintingNotationIllustration}
\end{figure}


As we have stated in the introduction, our video inpainting
algorithm takes a non-local patch-based approach. At the heart of such algorithms lies a global
patch-based functional which is to be optimised. This optimisation is
carried out using an iterative
algorithm inspired by the work of Wexler \emph{et al}. \cite{Wexler2007SpaceTime}.
The central machinery of the algorithm is based on the
alternation of two core steps: a search for the nearest neighbours of patches
which contain occluded pixels, and a reconstruction step based on the aggregation of the information
provided by the nearest neighbours.

This iterative algorithm is embedded in a multi-resolution pyramid scheme,
similarly to \cite{Drori2003Fragmentbased,Wexler2007SpaceTime}.
The multi-resolution scheme is vital for the correct reconstruction of
structures and moving objects in large occlusions. The precise details
concerning the multi-resolution scheme can be found in Section~\ref{subsec:multiResolutionDetails}.

A summary of the different steps in our algorithm can be seen in Figure~\ref{fig:inpaintingAlgoDiagram},
and the whole algorithm may be seen in Alg. \ref{algo:completeAlgo}.

\paragraph{Notation} Before describing our algorithm, let us first of all set down some notation. A diagram
which summarises this notation can be seen in Figure~\ref{fig:patchInpaintingNotationIllustration}.
Let $u:\Omega\rightarrow \mathbb{R}^3$ represent the colour video content, defined over a spatio-temporal volume $\Omega$.
In order to simplify the notation, $u$ will correspond both to the information
being reconstructed inside the occlusion, and the unoccluded information
which will be used for inpainting. We denote a spatio-temporal
position in the video as $p = (x,y,t)\in\Omega$ and by $u(p)\in\mathbb{R}^3$
the vector containing the colour values of the video at this position.

Let $\cH$ be the spatio-temporal occlusion (the ``hole'' to inpaint) and $\D$ the 
data set (the unoccluded area). Note that $\cH$ and $\D$
correspond to spatio-temporal
\emph{positions} rather than actual video content and they form a partition of $\Omega$, that is $\Omega=\cH\cup\D$ and $\cH\cap\D =\emptyset$.

Let $\N_p$ be a spatio-temporal
neighbourhood of $p$. This neighbourhood is defined as a rectangular cuboid centred on $p$.
The video \emph{patch} centered at $p$ is defined as vector $W_p^u = (u(q_1)\cdots u(q_N))$ of size $3\times N$, where the $N$ pixels in $\N_p$, $q_1\cdots q_N$, are ordered in a predefined way. 

Let us note $\tilde{\D}=\{p\in\D:\N_p\subset\D\}$ the
set of unoccluded pixels whose neighborhood is also unoccluded (video patch $W^u_p$ is only composed of known color values). 
We shall only use patches stemming from $\tilde{\D}$ to inpaint the occlusion. Also,
let $\tilde{\cH} = \cup_{p\in\cH}\N_p$ represent a dilated version of $\cH$.

Given a distance $d(\cdot,\cdot)$ between video patches, a key tool for patch-based inpainting is to define a correspondence map that associates to each pixel $p\in\Omega$ (notably those in occlusion) another position $q\in\tilde{\D}$, such that patches $W^u_p$ and $W^u_q$ are as similar as possible. This can be formalized using the so-called 
\emph{shift map} $\phi:\Omega\rightarrow \mathbb{R}^3$ that captures the shift between a position and its correspondent, that is $q=p+\phi(p)$ is the ``correspondent" of $p$.
This map must verify that $p+\phi(p)\in\tilde{\D},~\forall p$ (see Figure~\ref{fig:patchInpaintingNotationIllustration}
for an illustration).


\afterpage{
\begin{figure}[p]
\centering
\includegraphics[width = 0.7\linewidth]{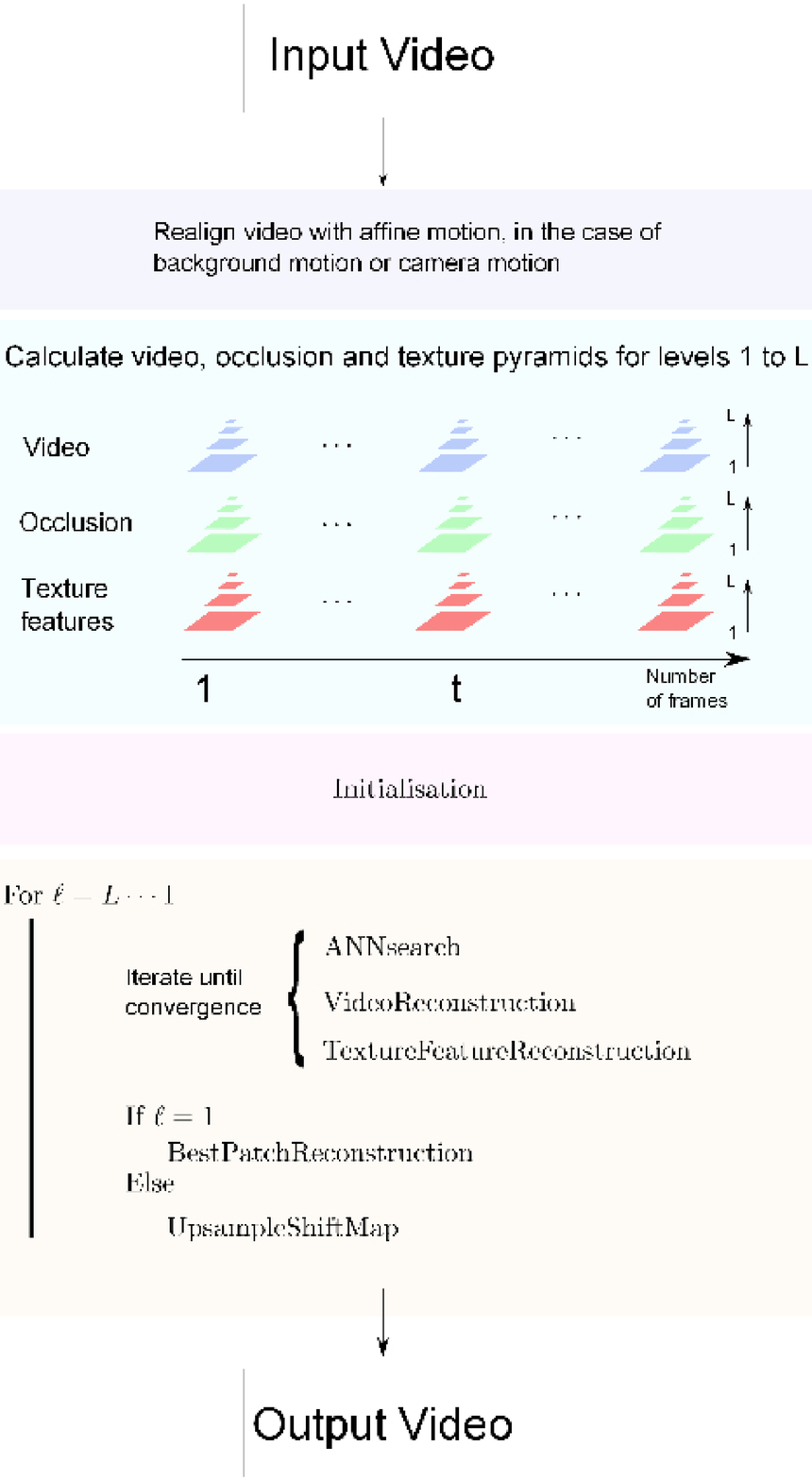}
\caption{Diagram of the proposed video inpainting algorithm.}
\label{fig:inpaintingAlgoDiagram}
\end{figure}
\clearpage
}

\paragraph{Mimizing a non-local patch-based functional}

The cost function which we use, following
the work of Wexler \emph{et al}. \cite{Wexler2007SpaceTime}, has both $u$ and $\phi$ as arguments:
\begin{equation}
\label{eq:wexlerInpaintingEnergy}
	E(u, \phi) = \sum_{p \in \cH} d^2(W_p^u,W_{p+\phi(p)}^u),
\end{equation}
with
\begin{equation}
\label{eq:patchDistance}
	d^2(W_p^u,W_{p+\phi(p)}^u) = \frac{1}{N} \sum_{q \in \N_p} \| u(q) - u(q+\phi(p))\|^2_2 .
\end{equation}

In all that follows, in order to avoid cumbersome notation we shall drop the $u$ from $W_p^u$ and
simply denote patches as $W_p$.

We show in Appendix~\ref{sec:variationalGraphCutsLink}, that this functional
is in fact a special case of the formulation of Arias \emph{et al}. \cite{Arias2011Variational}.
As mentioned at the beginning of this Section, this functional is optimised using the following two steps:

\begin{description}
\item[Matching] Given current video $u$, find in $\tilde{\D}$ the \textit{nearest neighbour} (NN) of each patch $W_p$ that has pixels in inpainting domain $\cH$, that is, the map $\phi(p),~\forall p \in \Omega\setminus\tilde{\D}$. 
\item[Reconstruction] Given shift map $\phi$, attribute a new value $u(p)$ to each pixel $p\in\cH$.
\end{description}

These steps are iterated
so as to converge to a satisfactory solution. The process may be seen
as an alternated minimisation of cost (\ref{eq:wexlerInpaintingEnergy}) over the shift map $\phi$ and the
video content $u$. 
As in many image processing and computer vision problems, this
approach is implemented in a multi-resolution framework in order
to improve results and avoid local minima.

\section{Proposed algorithm}
\label{sec:method}

Now that we have given a general outline of our
algorithm, we proceed to address some of the
key challenges in video inpainting. The first of these
concerns the search for the nearest neighbours of
patches centred on pixels which need to be inpainted.

\subsection{Approximate Nearest Neighbour (ANN) search}
\label{subsec:annSearch}

When considering the high complexity of the NN search step,
it quickly becomes apparent that searching for exact nearest neighbours
would take far too long. Therefore, an \emph{approximate nearest neighbour} (ANN)
search is carried out. Wexler \emph{et al}. proposed the k-d tree
based approach of Arya and Mount \cite{Arya1993Approximate} for this step,
but this approach remains quite slow.
For example, one ANN search step takes about an hour for a video containing $120 \times 340 \times 100$ pixels, with
about $422,000$ missing pixels, which
represents a relatively small occlusion (the equivalent of
a $65 \times 65$ pixel box in each frame).
We shall address this problem here, in particular by using
an extension of the PatchMatch algorithm \cite{Barnes2009PatchMatch}
to the spatio-temporal case. We note that the PatchMatch algorithm
has also been used in conjunction with a 2D version of Wexler's
algorithm for \emph{image} inpainting, in the Content-Aware Fill tool of
Photoshop \cite{PhotoshopCS5}, and by Darabi \emph{et al}. \cite{Darabi2012Image}.


Barnes \etal 's PatchMatch 
is a conceptually simple algorithm based
on the hypothesis that, in the case of image patches,
the shift map defined by the spatial offsets between ANNs is piece-wise constant.
This is essentially because the image elements which
the ANNs connect are often on rigid objects
of a certain size. In essence, the algorithm looks
randomly for ANNs and tries to ``spread'' those
which are good.
We extend this principle to the spatio-temporal
setting.
Our spatio-temporal extension of the PatchMatch algorithm consists of three steps:
(i) initialisation,(ii) propagation and (iii) random search.

Let us recall that $\tilde{\cH}$ is a dilated version of $\cH$.
Initialisation consists of randomly associating
an ANN to each patch $W_p$, $p\in\tilde{\cH}$, which
gives an initial ANN shift map, $\phi$.
In fact, apart from the first iteration, we already have a good initialisation:
the shift map $\phi$ from the previous iteration. Therefore,
except during the initialisation step (see Section~\ref{subsec:inpaintingInitialisation}),
we use this previous shift map in our algorithm instead of initialising randomly.

The propagation step encourages shifts in $\phi$
which lead to good ANNs to be spread throughout $\phi$.
In this step, 
all positions in the video volume are scanned lexicographically.
For a given patch $W_p$ at location $p=(x,y,t)$, the algorithm considers the
following three candidates : 
$W_{p + \phi(x-1,y,t)}$, $W_{p + \phi(x,y-1,t)}$ and $W_{p + \phi(x,y,t-1)}$.
If one of these three patches 
has a smaller patch distance with respect to $W_p$ than
$W_{p + \phi(p)}$, then $\phi(p)$ is replaced with the new, better shift.
The scanning order is reversed for the next iteration of the
propagation, and the algorithm tests
$W_{p + \phi(x+1,y,t)}$, $W_{(p + \phi(x,y+1,t)}$ and $W_{p + \phi(x,y,t+1)}$.
In the two different scanning orderings, the important point
is obviously to use the patches which have already been processed
in the current propagation step.

The third step, the random search, consists in
looking randomly for better ANNs of each $W_p$ in an increasingly
small area around $p + \phi(p)$, starting with a
maximum search distance. At iteration $k$, the random candidates are centred at the
following positions:
\begin{equation}
\label{eq:rechercheAleatoire}
q = p + \phi(p) + \lfloor r_{\mathrm{max}}\rho^k \delta_k\rfloor,
\end{equation}
where $r_{\max}$ is the maximum search radius around $p + \phi(p)$, $\delta_k$ is
a 3-dimensional vector drawn from the uniform distribution over unit cube $[-1,1] \times [-1,1] \times [-1,1]$
and $\rho\in(0,1)$ is the reduction factor of the search window size.
In the original PatchMatch, $\rho$ is set to 0.5.
This random search avoids the algorithm getting stuck
in local minima. The maximum search parameter $r_{\mathrm{max}}$
is set to the maximum dimension of the video, at the current resolution level.

The propagation and random search steps are iterated several
times to converge to a good solution. In our work, we set this
number of iterations to 10.
For further details concerning the PatchMatch algorithm in the 2D case,
see \cite{Barnes2009PatchMatch}. Our spatio-temporal extension is summarized in Algorithm \ref{algo:ANNsearch}.

\begin{algorithm}
\SetAlgoLined
\KwData{Current inpainting configuration $u$, $\phi$, $\tilde{\cH}$}
 \KwResult{ANN shift map $\phi$}
 \BlankLine
\For{$k=1$ \KwTo $10$}
{
 \eIf(\tcc*[f]{Propagation on even iteration}){$k$ is even}
 {
 \For{$p=p_1$\KwTo$p_{|\tilde{\cH}|}$ (pixels in $\tilde{\cH}$ lexicographically ordered)} 
 {
 	$a=p-(1,0,0)$, $b=p-(0,1,0)$, $c=p-(0,0,1)$\; 
 	$q=\arg\min_{r\in\{p,a,b,c\}} d(W^u_p,W^u_{p+\phi(r)})$\;
 	\lIf{$p+\phi(q)\in\tilde{\D}$}{$\phi(p)\leftarrow\phi(q)$}
 }
 }(\tcc*[f]{Propagation on odd iteration})
 {
 \For{$p=p_{|\tilde{\cH}|}$ \KwTo $p_1$}
 {
 	$a=p+(1,0,0)$, $b=p+(0,1,0)$, $c=p+(0,0,1)$\; 
 	$q=\arg\min_{r\in\{p,a,b,c\}} d(W^u_p,W^u_{p+\phi(r)})$\;
 	\lIf{$p+\phi(q)\in\tilde{\D}$}{$\phi(p)\leftarrow\phi(q)$}
 }
 }
\For(\tcc*[f]{Random search}){$p=p_1$ \KwTo $p_{|\tilde{\cH}|}$}
 {
 	$q=p+\phi(p)+\lfloor r_{\max}\rho^k$RandUniform($[-1,1]^3$) $\rfloor$\; 
 	\lIf{$d(W^u_p,W^u_{p+\phi(q)})<d(W^u_p,W^u_{p+\phi(p)})$ {\textbf{and}} $p+\phi(q)\in\tilde{\D}$}{$\phi(p)\leftarrow\phi(q)$}
 } 
 }
 \caption{ANN search with 3D PatchMatch}
 \label{algo:ANNsearch}
\end{algorithm}

We note here that other ANN search methods for image patches exist
which outperform PatchMatch \cite{He2012Computing,Korman2011Coherency,Olonetsky2012TreeCANN}.
However, in practice, PatchMatch appeared to be a
good option because of its conceptual simplicity
and nonetheless very good performance.
Furthermore, to take the example of the ``TreeCANN'' method of Olonetsky and Avidan \cite{Olonetsky2012TreeCANN},
the reported reduction in execution time is largely based on a very good ANN shift map initialisation followed
by a small number of propagation steps. In our case, we already have a good initialisation
(from the previous iteration), which makes the usefulness of such approaches questionable.
However, further acceleration is certainly something which could be developed in
the future.

\subsection{Video reconstruction}
\label{subsec:videoReconstruction}

\begin{figure}
\begin{tabular}{c c}
\includegraphics[width = 0.5\linewidth]{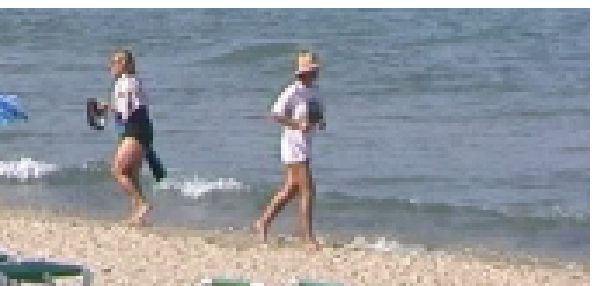} &
\includegraphics[width = 0.5\linewidth]{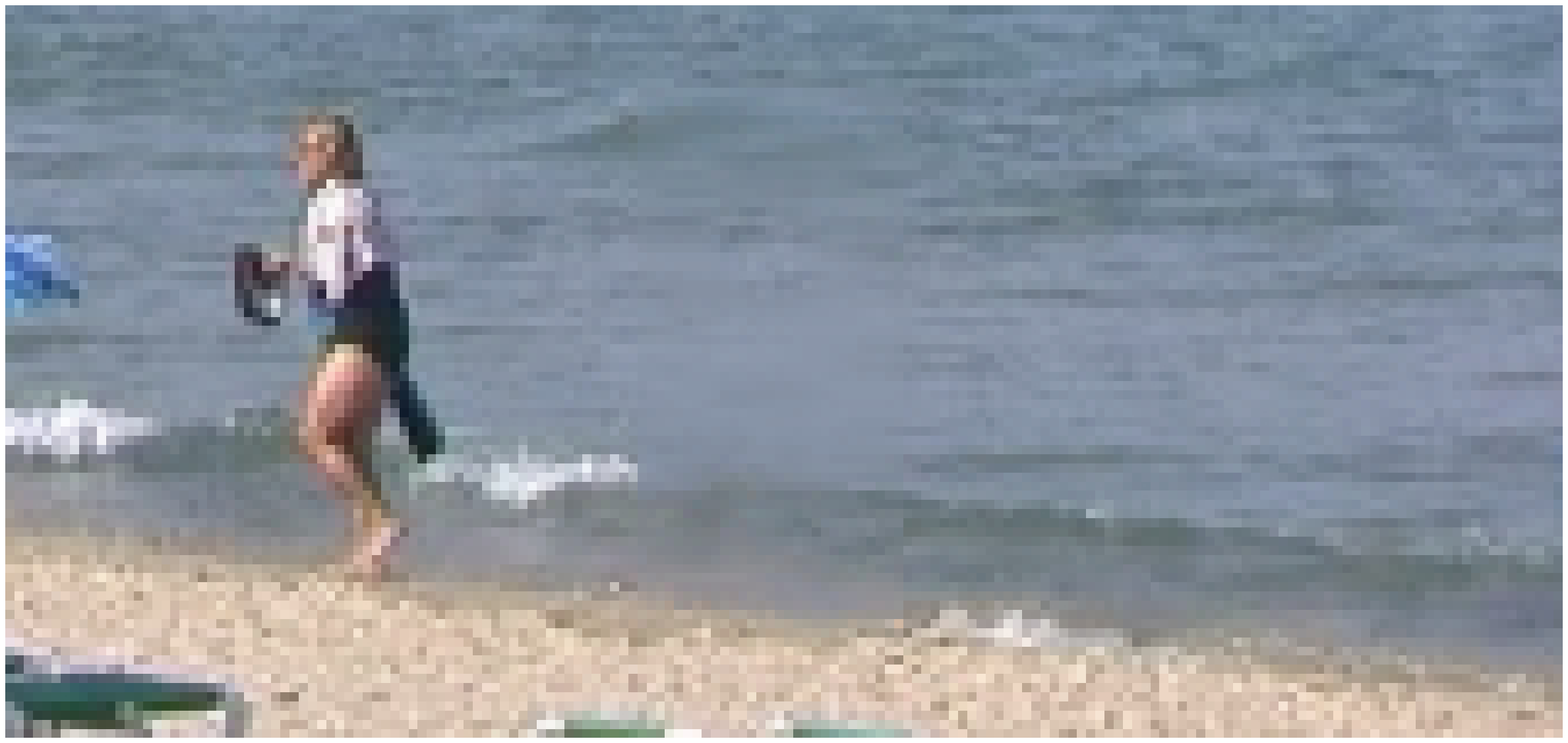} \\
\centering{Original frame : ``Crossing ladies'' \cite{Wexler2007SpaceTime}} & Weighted average \\
& \\
\includegraphics[width = 0.5\linewidth]{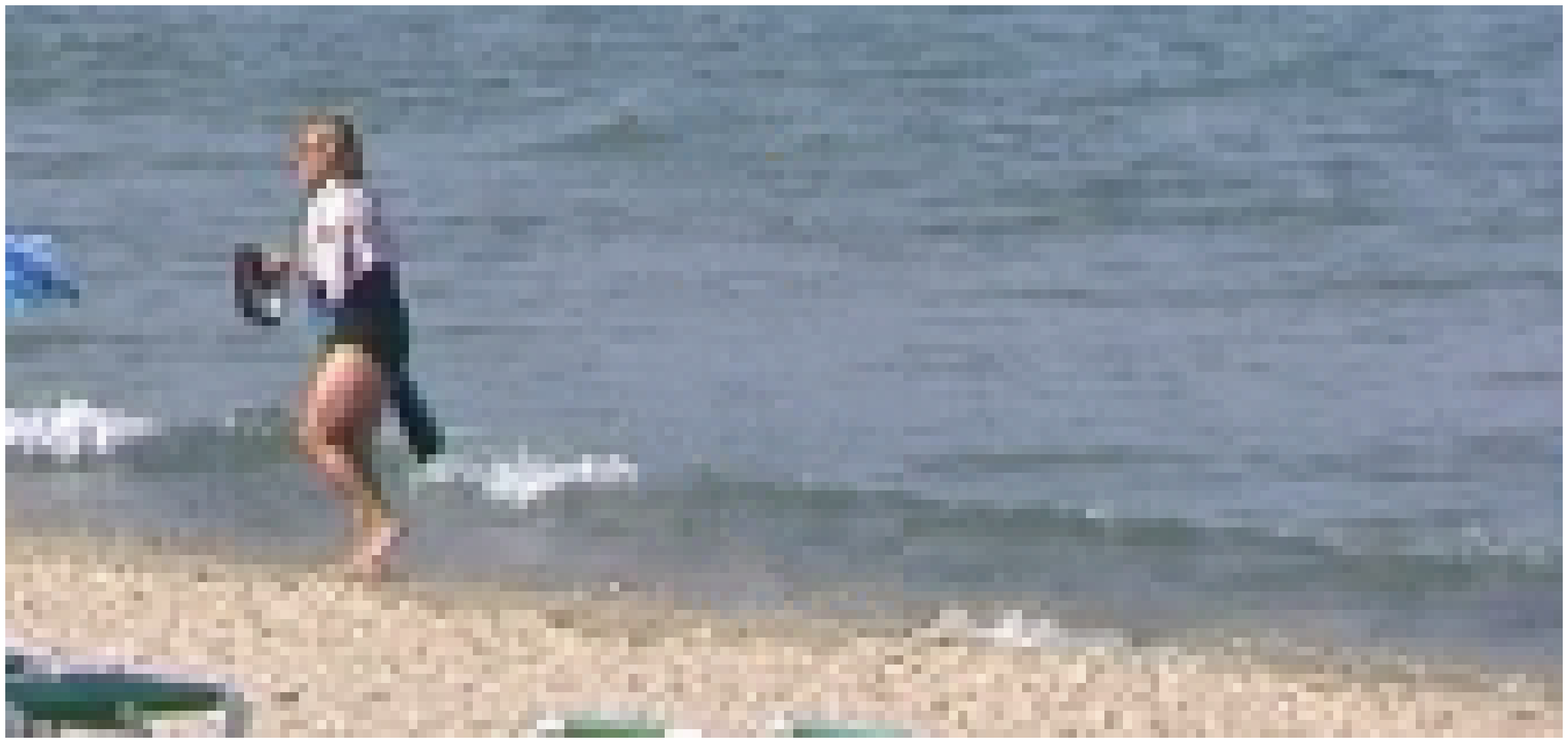} &
\includegraphics[width = 0.5\linewidth]{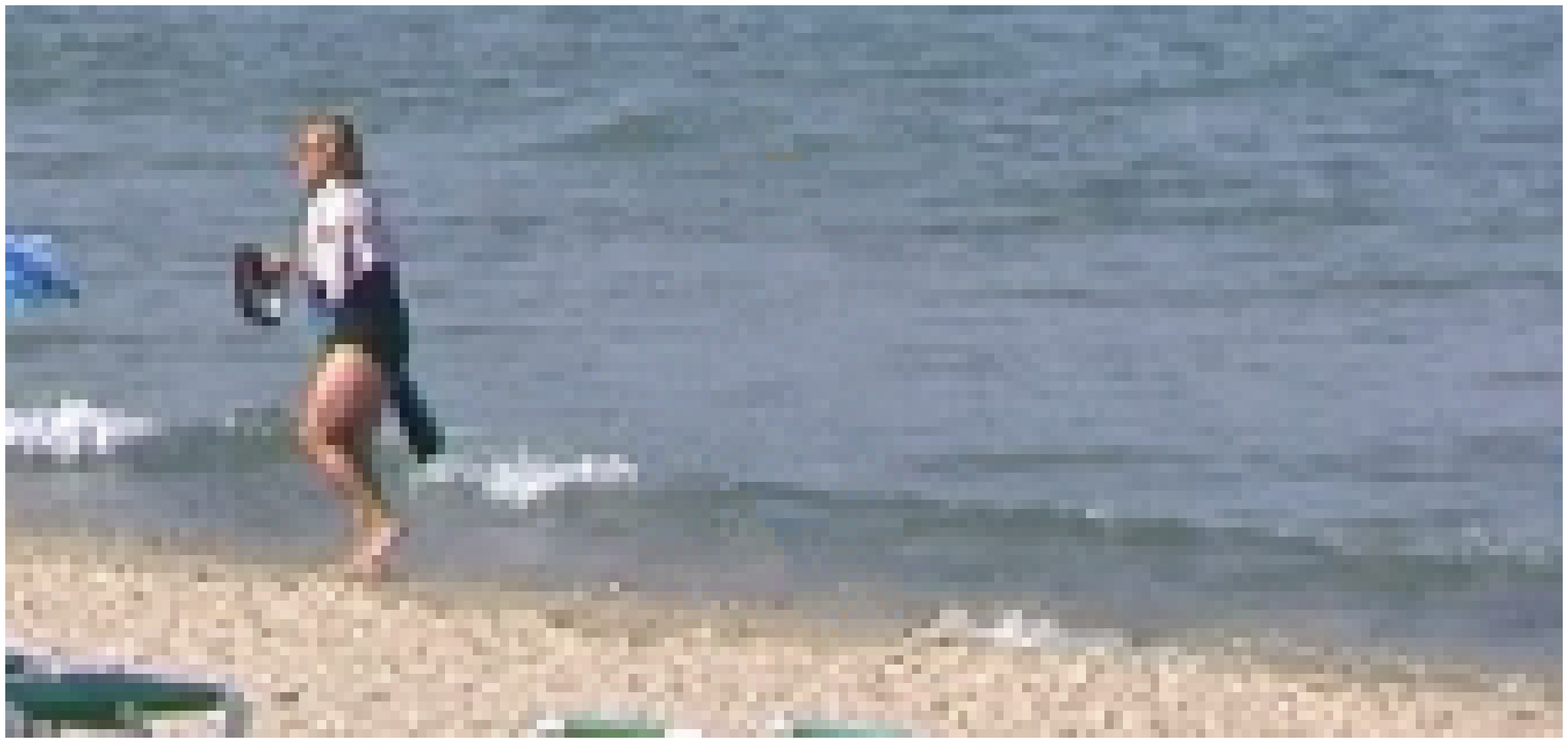} \\
\centering{Mean shift} & Best patch \\
\end{tabular}
\caption{{\bf Comparison of different final reconstruction methods}. We observe that the
proposed reconstruction using only the best patch at the end of the algorithm produces
similar results to the use of the mean shift algorithm, avoiding blur induced by weighted patch averaging, while being less computationally expensive. Please note that the blurring effect is best viewed in the pdf version of the paper.}
\label{fig:meanShiftComparison}
\end{figure}

Concerning the reconstruction step, we use a
a weighted mean based approach, inspired by the work of Wexler \emph{et al}., in which
each pixel is reconstructed in the following manner:
\begin{equation}
u(p) = \frac{\sum_{q \in \N_p} s_p^q \, u(p + \phi(q))}{\sum_{q \in \N_p} s_p^q},\quad\forall p\in\cH,
\label{eq:wexlerWeightedMean}
\end{equation}
\noindent with
\begin{equation}
s_p^q = \exp{ \left(-\frac{d^2(W_q,W_{q+\phi(q)})}{2 \sigma_p ^2} \right) }.
\label{eq:wexlerWeights}
\end{equation}

Wexler \emph{et al}. proposed the use of an additional weighting term 
to give more weight to the information near the occlusion border. We dispense
with this term, since in our scheme it is somewhat replaced by
our method of initialising the solution which will be detailed
in Section~\ref{subsec:inpaintingInitialisation}.
Parameter $\sigma_p$ is defined as the 75th percentile of all
distances $\{d(W_q,W_{q+\phi(q)}),~q \in \N_p\}$ as in \cite{Wexler2007SpaceTime}.

Observe that in order to minimise (\ref{eq:wexlerInpaintingEnergy}), the natural
approach would be to do the reconstruction with the non-weighted scheme ($s_p^q=1$ in 
Equation \ref{eq:wexlerWeightedMean}) that stems from $\frac{\partial E}{\partial u(p)}=0$. 
However, the weighted scheme above 
tends to accelerate the convergence of the algorithm, meaning that we produce
good results faster.


An important observation is that, in the case of regions with high
frequency details, the use of this mean reconstruction (weighted or unweighted) often
leads to blurry results, even if the correct patches have been
identified. This phenomenon was also noted in \cite{Wexler2007SpaceTime}. 
Although we shall propose in Section~\ref{subsec:inpaintingTextures} a method
to correctly identify textured patches in the matching steps, this does not deal with the \emph{reconstruction} of the video. 
We thus need to address this problem, at least in the final stage of the approach:
throughout the algorithm, we use the unweighted mean
given in Equation~\ref{eq:wexlerWeightedMean} and, at the end of the algorithm, when the solution has converged
at the finest pyramid level, we simply inpaint the occlusion using the best patch among the contributing ones.
This corresponds to setting $\sigma_p$ to 0 in (\ref{eq:wexlerWeightedMean}-\ref{eq:wexlerWeights})
or, seen in another light, may be viewed as a very crude annealing procedure. Final reconstruction at position $p\in\cH$ reads:
\begin{equation}
u^{\mathrm{(final)}}(p) = u(p + \phi^{\mathrm{(final)}}(q^*)),\quad\mathrm{with~}q^*=\arg\min_{q\in\N_p}d(W^{(\mathrm{final}-1)}_q,W_{q+\phi^{\mathrm{(final)}}(q)}).
\label{eq:finalReconstruction}
\end{equation}
Another solution to this problem based on the mean shift algorithm was
proposed by Wexler \emph{et al}. in \cite{Wexler2007SpaceTime}, but such an approach increases the complexity
and execution time of the algorithm.
Figure~\ref{fig:meanShiftComparison} shows that very similar 
results to those in \cite{Wexler2007SpaceTime} may be obtained with 
our much simpler approach.

\subsection{Video texture pyramid}
\label{subsec:inpaintingTextures}



\begin{figure}
\begin{tabularx}{\linewidth}{@{} *2{>{\centering\arraybackslash}X}@{}}
\includegraphics[width = 0.8\linewidth]{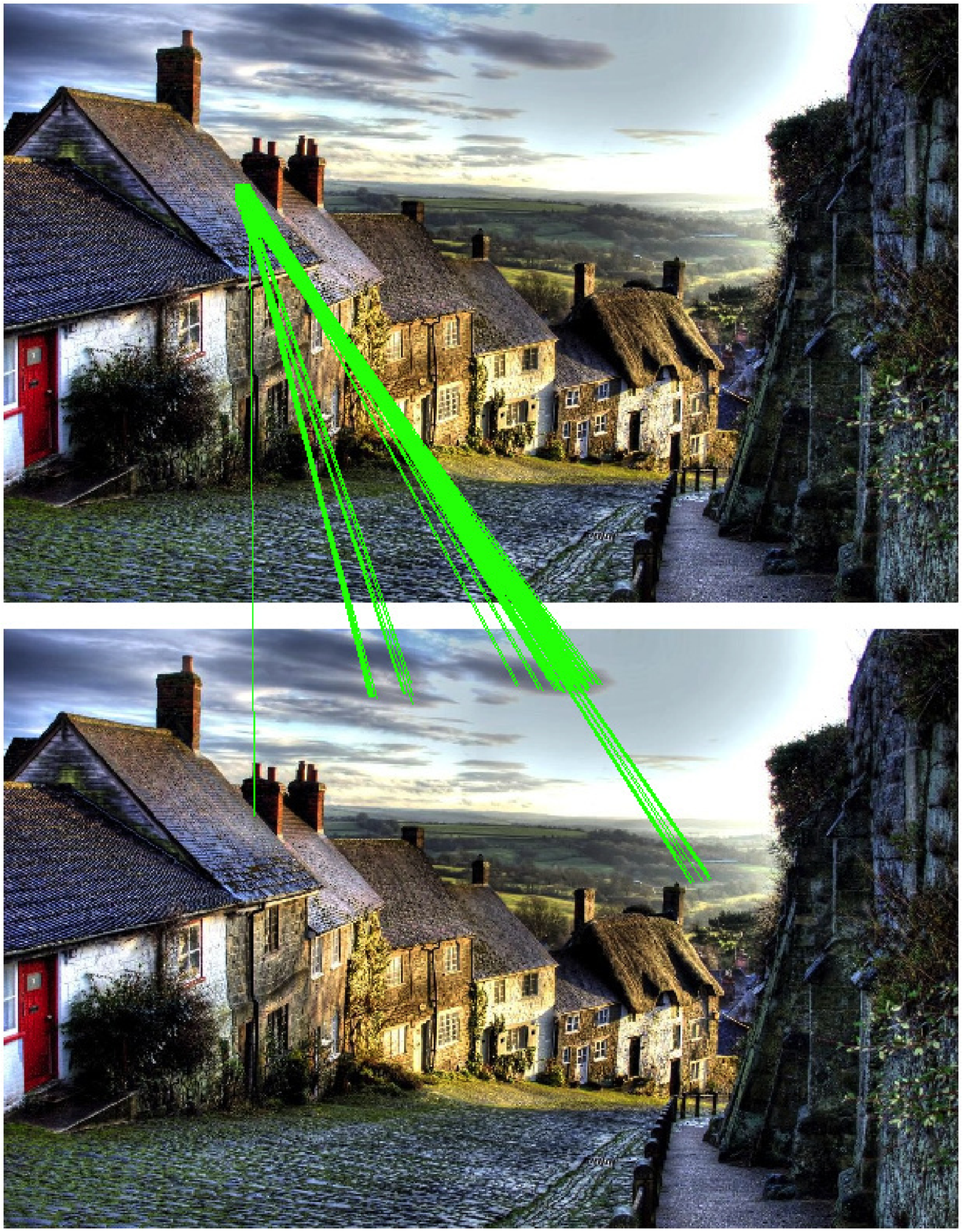}
&
\includegraphics[width = 0.8\linewidth]{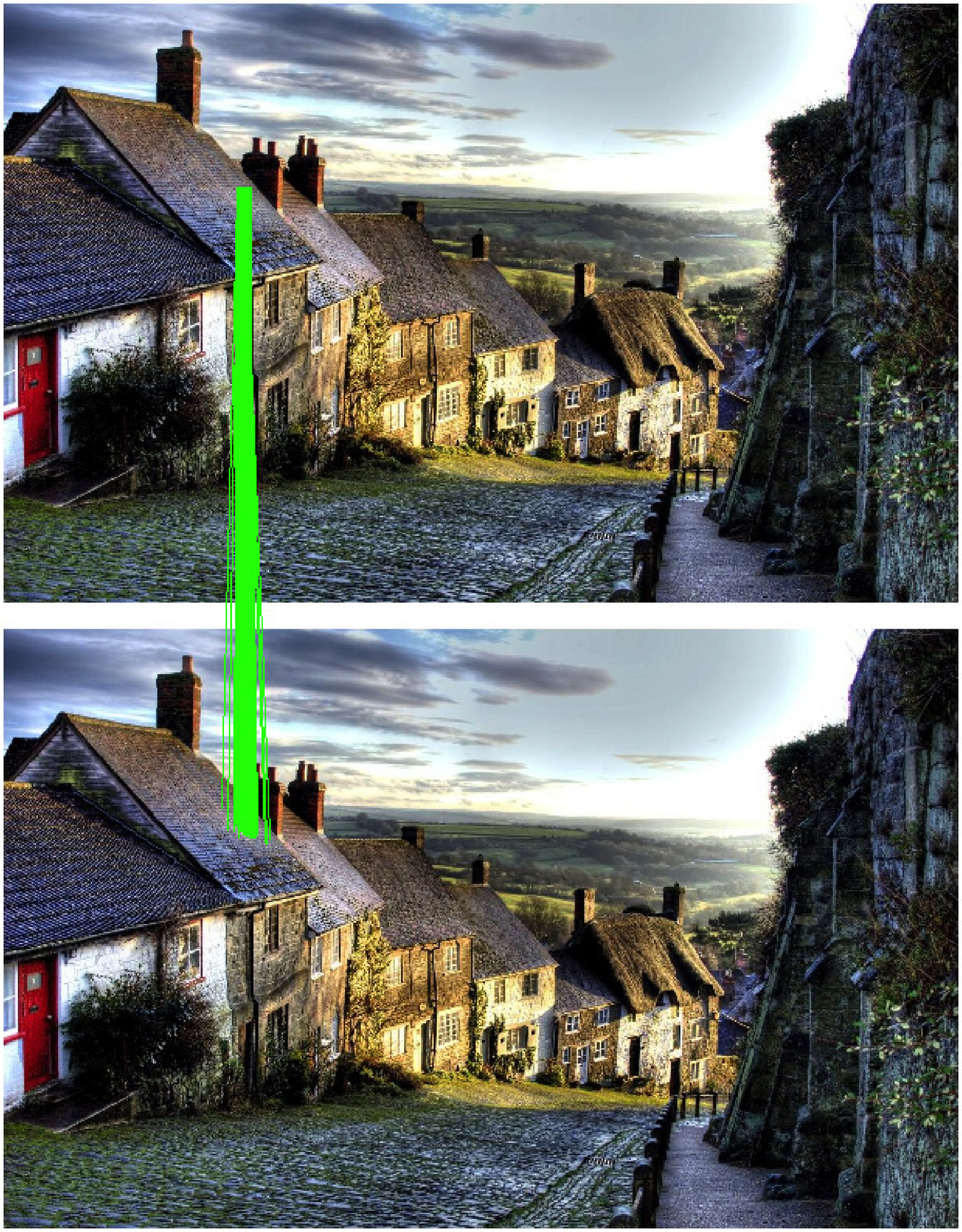}
\\
Approximate nearest neighbours without texture features &
Approximate nearest neighbours with texture features \\
\end{tabularx}
\caption{{\bf Illustration of the necessity of texture features for inpainting}. Without the texture features,
the correct texture may not be found.}
\label{fig:americanFeatures}
\end{figure}


In order for any patch-based inpainting algorithm to
work, it is necessary that the patch distance identify
``correct'' patches. This is not
the case in several situations. Firstly, as
noticed by Liu and Caselles in \cite{Liu2013ExemplarBased},
the use of multi-resolution pyramids can make patch comparisons
ambiguous, especially in the case of textures, in
images and videos. Secondly, it turns out that the commonly
used $\ell^2$ patch distance is ill-adapted to comparing
textured patches. Thirdly, PatchMatch itself can contribute
to the identification of incorrect patches. These reasons
are explored and explained more extensively in Appendix~\ref{sec:texturedPatchesAnalysis}.
A visual illustration of the problem may be seen in Figure~\ref{fig:americanFeatures}.
We note here that similar ambiguities were also identified by Bugeau \emph{et al}.
in \cite{Bugeau2010Comprehensive}, but their interpretation
of the phenomenon was somewhat different.

We propose a solution to this problem in the form of
a multi-resolution pyramid which reflects the textural
nature of the video. We shall refer to this as \textit{the texture
feature pyramid}. The information in this texture feature
pyramid is added to the patch distance in order to identify
the correct patches.


In order to identify textures, we shall consider a simple, gradient-based texture attribute.
Following Liu and Caselles \cite{Liu2013ExemplarBased},
we consider the absolute value of the image derivatives,
averaged over a certain spatial neighbourhood $\nu$.
Obviously, many other attributes could be considered, however
they seemed too involved for our purposes.

More formally, we introduce the two-dimensional texture feature $T = (T_x,T_y)$,
computed at each pixel $p\in\Omega$:
\begin{equation}
T(p)  = \frac{1}{\mathrm{card}(\nu)}\sum_{q \in \nu} (|I_x(q)|,|I_y(q)|),
\label{eq:textureFeature}
\end{equation}
where $I_x(q)$ (resp. $I_y(q)$) is the derivative of the image intensity (grey-level)
in the $x$ (resp. $y$) direction at the pixel $q$.
The squared patch distance is now defined as
\begin{equation}
	d^2(W_p,W_q) = \frac{1}{N} \sum_{r \in \N_p} \Big(\|u(r) - u(r-p+q))\|^2_2 + \lambda \|T(r)-T(r-p+q)\|_2^2\Big),
\end{equation}
where $\lambda$ is a weighting scalar.

The feature pyramid is then set up by subsampling the
texture features of the full-resolution input video $u$. We note here that each level is obtained
by subsampling the information contained at the \emph{finest pyramid resolution},
and \emph{not} by calculating $T^{\ell}$ based on the subsampled
video at the level $\ell$:
\begin{equation}
\forall (x,y,t)\in \Omega^{\ell}, ~T^{\ell}(x,y,t)=T(2^{\ell}x,2^{\ell}y,t),\quad \ell=1\cdots L.
\label{eq:texturePyr}
\end{equation}
This is an important point, since the required textural information \emph{does not exist}
at coarser levels. Features are not filtered before subsampling,
since they have already been averaged over the neighbourhood $\nu$.
In the experiments done in this paper, this neighbourhood is set, by default, to the area to which a coarsest-level pixel corresponds, which is
a square of size $2^{L-1}$, as is done in \cite{Liu2013ExemplarBased}. However, in a more
general setting, the size of this
area should be independent of the number of levels,
so care should be taken in the case where
few pyramid levels are used.

%
%
%

A notable difference
with respect to the work of Liu and Caselles \cite{Liu2013ExemplarBased}
is the fact that we use the texture features at \emph{all} pyramid levels.
Liu and Caselles do not do this, since they perform graph cut based
optimisation at the coarsest level, and at the finer levels only consider small relative
shifts with respect to the coarse solution.

A final choice which must be made when using the texture features
is how they are \emph{themselves} reconstructed. In shift maps based
algorithms, this is not a problem, since by definition an occluded pixel
takes on the characteristics of its correspondent in $\D$ 
(colour, texture features or anything else).

In our case, we inpaint the texture features using the
same reconstruction scheme as is used for colour information (see Eq.~\ref{eq:wexlerWeightedMean}):

\begin{equation}
T(p) = \frac{\sum_{q \in \N_p} s_p^q T(p + \phi(q))}{\sum_{q \in \N_p} s_p^q},~\forall p\in\cH.
\label{eq:textureFeatureReconstruction}
\end{equation}

Conceptually, the use of these features is quite simple, and easily fits into our inpainting
framework.
To summarise, these features may be seen as simple texture descriptors which help the
algorithm avoid making mistakes when choosing the area to use for inpainting.

The methodology which we have proposed for dealing with
dynamic video textures is important for the following reasons. Firstly,
to the best of our knowledge, this is the first inpainting approach which proposes
a global optimisation and which can deal correctly with textures in images and videos,
without restricting the search space (contrary to
\cite{Criminisi2003Object,Darabi2012Image,Granados2012How,Liu2013ExemplarBased,Pritch2009Shiftmap}).
Secondly,
while the problem of recreating video textures is a research subject
in its own right and algorithms have been developed for their
synthesis \cite{Doretto2003Dynamic,Kwatra2003Graphcut,Schodl2000Video}, ours
is the first algorithm to achieve this within an inpainting framework.
Finally we note that algorithms such as that of Granados \emph{et al}. \cite{Granados2012Background},
which are specifically dedicated to background reconstruction, cannot
deal with background video textures (such as waves), since they suppose
that the background is rigid. This hypothesis is clearly not true for
video textures. An example of the impact of the texture features in
video inpainting may be seen in Figure~\ref{fig:videoTexture}.

\subsection{Inpainting with mobile background}
\label{subsec:movingBackgroundInpainting}

\begin{figure}[t]
\begin{tabularx}{\linewidth}{@{} *3{>{\centering\arraybackslash}X}@{}}
\includegraphics[width = \linewidth]{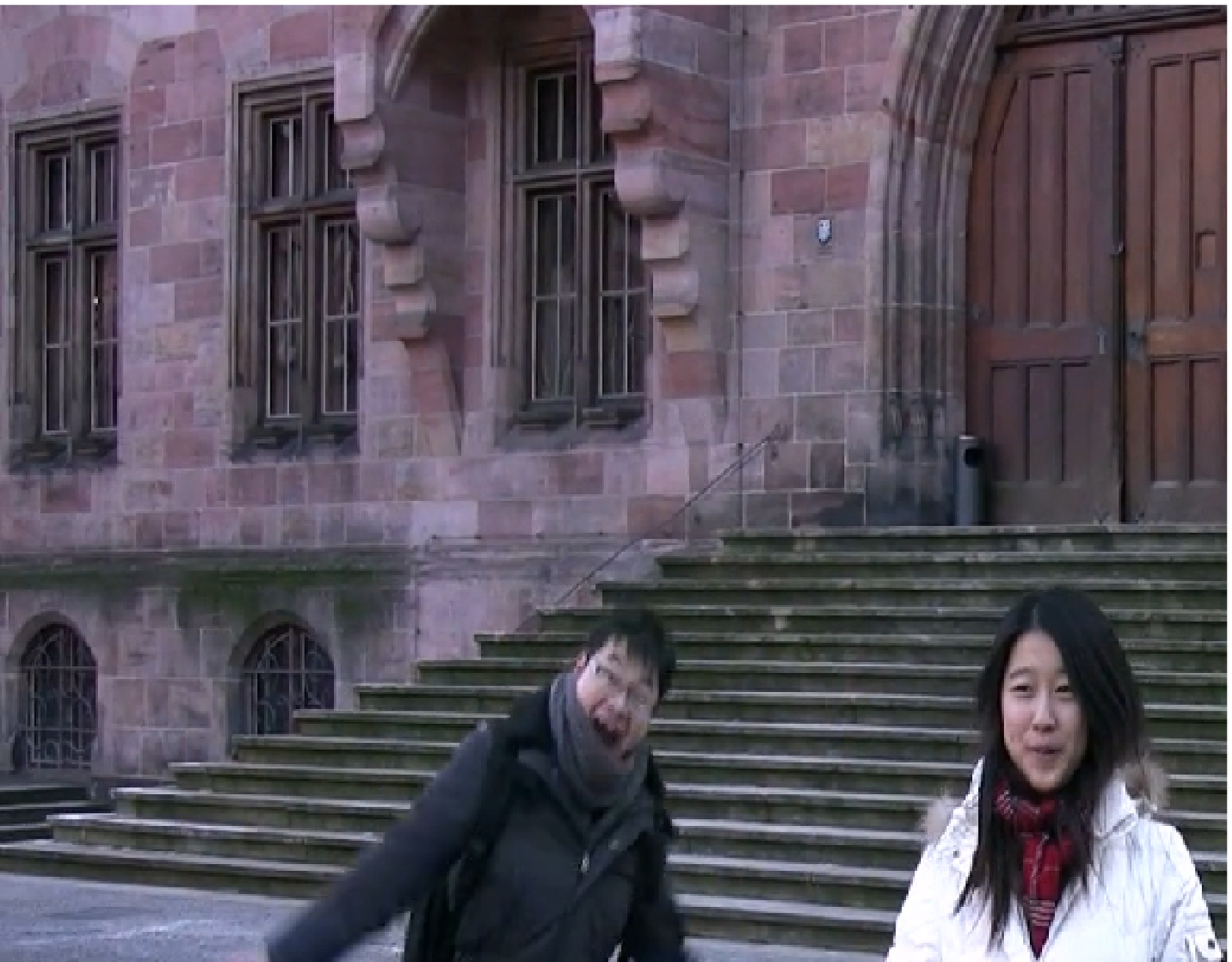} &
\includegraphics[width = \linewidth]{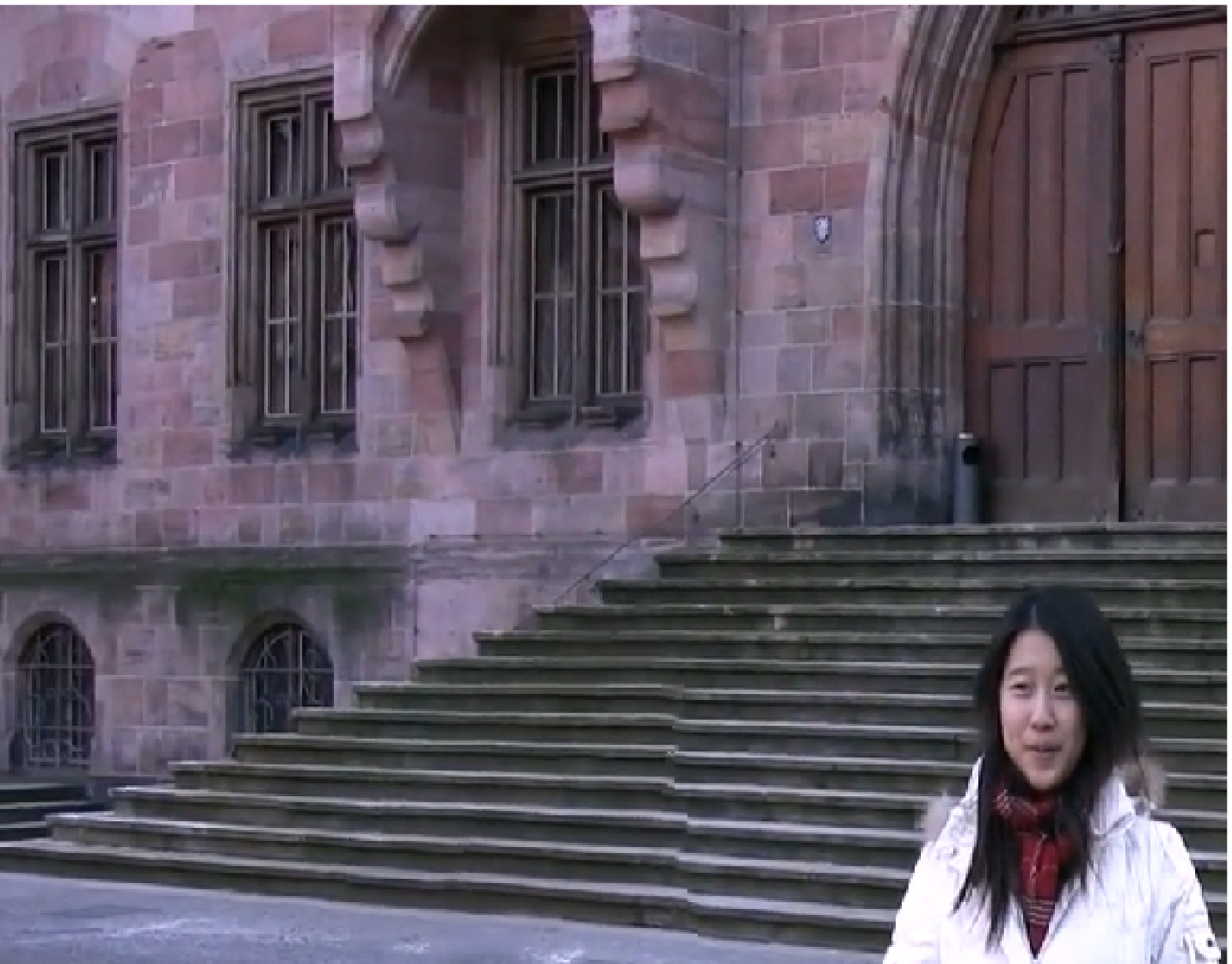} &
\includegraphics[width = \linewidth]{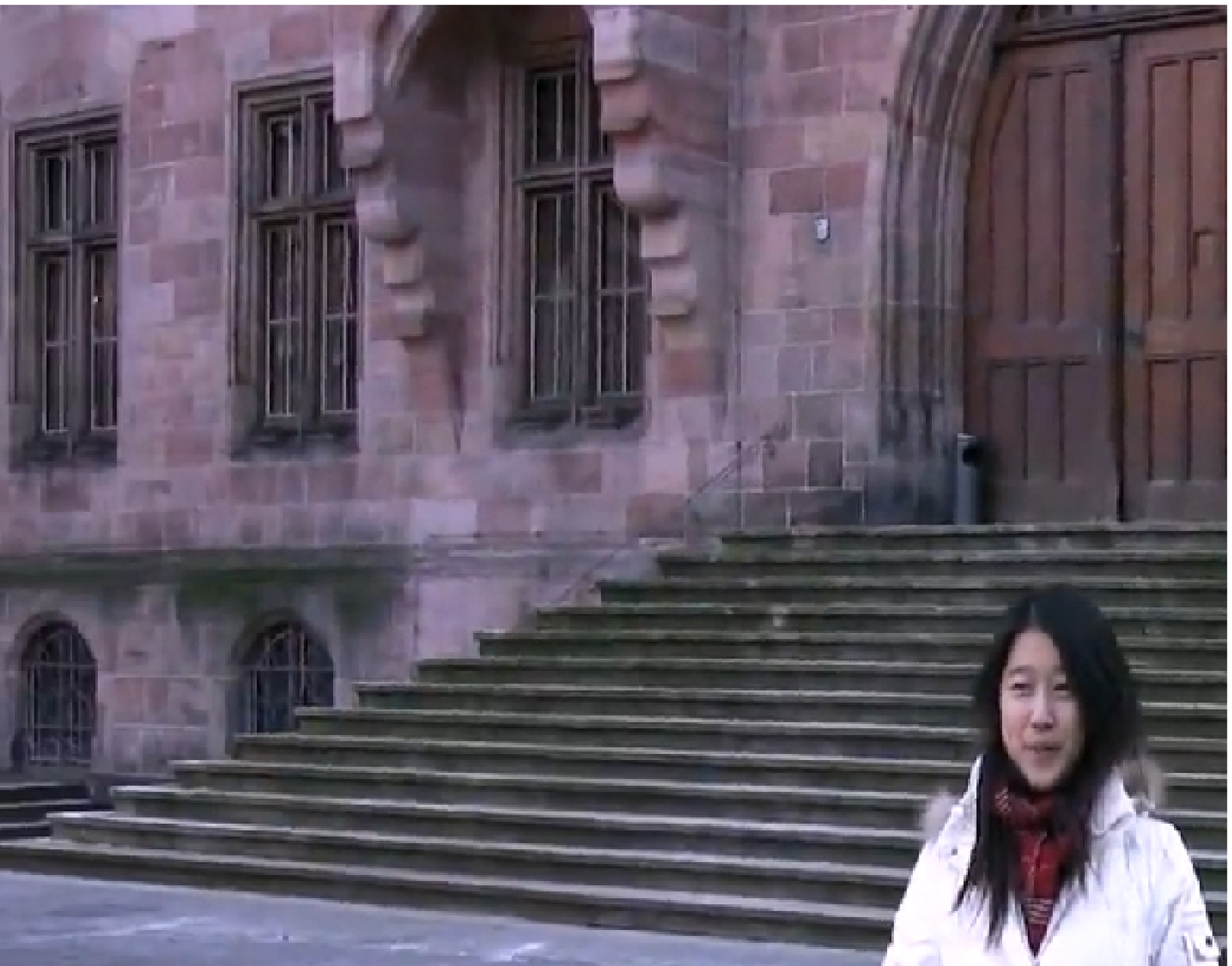} \\
\centering{Original frame : ``Jumping man'' \cite{Granados2012Background}}
& Inpainting without realignement & Inpainting with realignement\\
& & \\
\end{tabularx}
\caption{{\bf A comparison of inpainting results with and without affine realignment}.
		Notice the
		incoherent reconstruction on the steps, due to random camera
		motion which makes spatio-temporal patches difficult to compare. These random
		motions are corrected with affine motion estimation and inpainting is performed in the
		realigned domain.}
\label{fig:withWithoutRealignement}
\end{figure}


We now turn to another common case of video inpainting, that of
mobile backgrounds. This is the case, for example, when hand-held
cameras are used to capture the input video.

There are several possible solutions to this problem.
Patwardhan \emph{et al}. \cite{Patwardhan2007Video} segment the video into moving foreground
and background (which may also display motion) using
motion estimation with block matching. Once the
moving foreground is inpainted, the background is
realigned with respect to a motion estimated with block matching, and the background is filled by copying and
pasting background pixels. In this case, the background should be perfectly realigned.

Granados \emph{et al}. \cite{Granados2012Background} propose
a homography-based algorithm for this task. They estimate a set of
homographies between each frame and choose which homography
should be used for each occluded pixel belonging to the background.

Both of these algorithms require that the background and
foreground be segmented, which we wish to avoid here. Furthermore, they have the quite
strict limitation that pixels are simply copied from their
realigned positions, meaning
that the realignment must be extremely accurate.
Here, we propose a solution which allows us to use our patch-based variational
framework for both tasks (foreground and background inpainting) simultaneously, without any segmentation
of the video into foreground and background.

\begin{algorithm}
\SetAlgoLined
\KwData{Input video $u$}
\KwResult{Aligned video and affine warps}
\BlankLine
$N_f = $ number of frames in video\;
$N_m = \lfloor \frac{N_f}{2} \rfloor $\;
\For{$n=1$ \KwTo $N_f-1$} 
{
	$\theta_{n,n+1}	\leftarrow$ EstimateAffineMotion($u_n,u_{n+1}$)\;
}
\For{$n=1$ \KwTo $N_f$} 
{
	\lIf{$n<N_m$}{$\theta_{n,N_m}=\theta_{N_m-1,N_m}\circ\cdots\circ\theta_{n,n+1}$}
	\lIf{$n>N_m$}{$\theta_{n,N_m}=\theta^{-1}_{N_m-1,N_m}\circ\cdots\circ\theta^{-1}_{n,n+1}$}
	$u_n \leftarrow $ AffineWrap($u_n,\theta_{n,N_m}$)\;
}
\caption{Pre-processing step for realigning the input video.}
\label{algo:realignmentPreProcessing}
\end{algorithm}

The fundamental hypothesis behind patch-based methods in images or
videos is that content is redundant and repetitive. This is
easy to see in images, and may appear to be the case in videos.
However, the temporal dimension is added in video patches,
meaning that a \emph{sequence} of image patches should be repeated
throughout the video. This is not the case when a video displays
random motion (as with a mobile camera): even if the required content
appears at some point in the sequence, there is no guarantee that
the required spatio-temporal patches will repeat themselves
with the same motion. Empirically, we have observed that
this is a significant problem even in the case of motions with
small amplitude.

To counter this problem, we estimate a dominant, affine motion between each pair of successive frames,
and use this to realign each frame with respect to one reference
frame. In our work, we chose the reference frame to
be the middle frame of the sequence (this should be adapted for larger sequences). We use the work of
Odobez and Bouthemy \cite{Odobez1995Robust} to realign
the frames.  The colour values of the pixels in the realigned frames are obtained
using linear interpolation. The occlusion $\cH$
is obviously also realigned.
Once the frames are realigned with the reference frame (Alg. \ref{algo:realignmentPreProcessing}),
we inpaint the video as usual. Finally, when inpainting
is finished, we perform the inverse affine transformation on the images and paste the
solution into the original occluded area.
Figure~\ref{fig:withWithoutRealignement} compares the results with and without
this pre-processing step, on a prototypical example. Without it, it is not possible to find coherent
patches which respect the border conditions.

\subsection{Initialisation of the solution}
\label{subsec:inpaintingInitialisation}


\begin{figure}[t]
\begin{tabularx}{\linewidth}{X X X X}
\includegraphics[width = \linewidth]{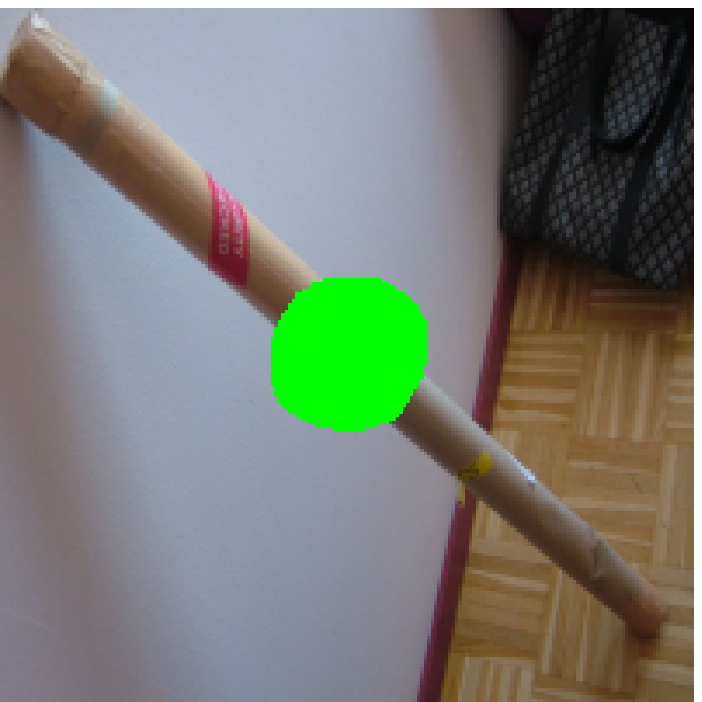} &
\includegraphics[width = \linewidth]{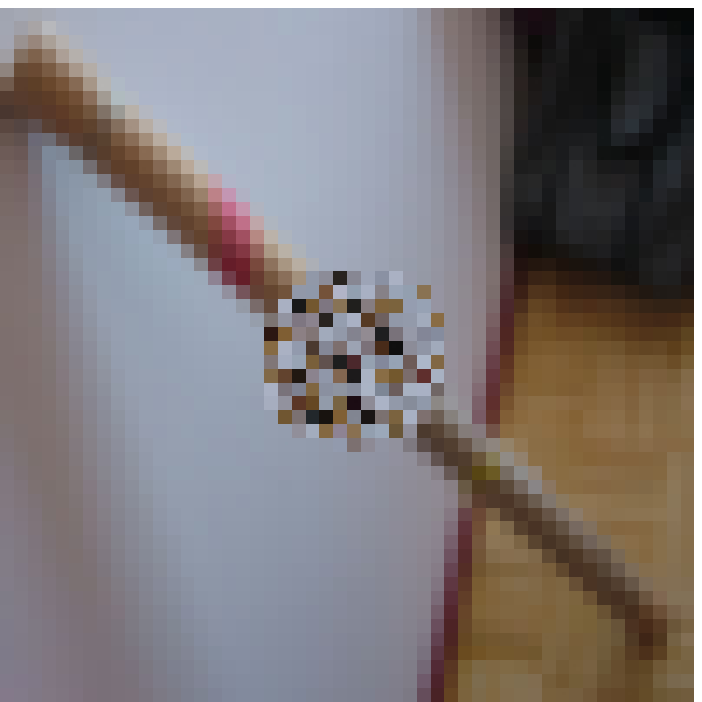} &
\includegraphics[width = \linewidth]{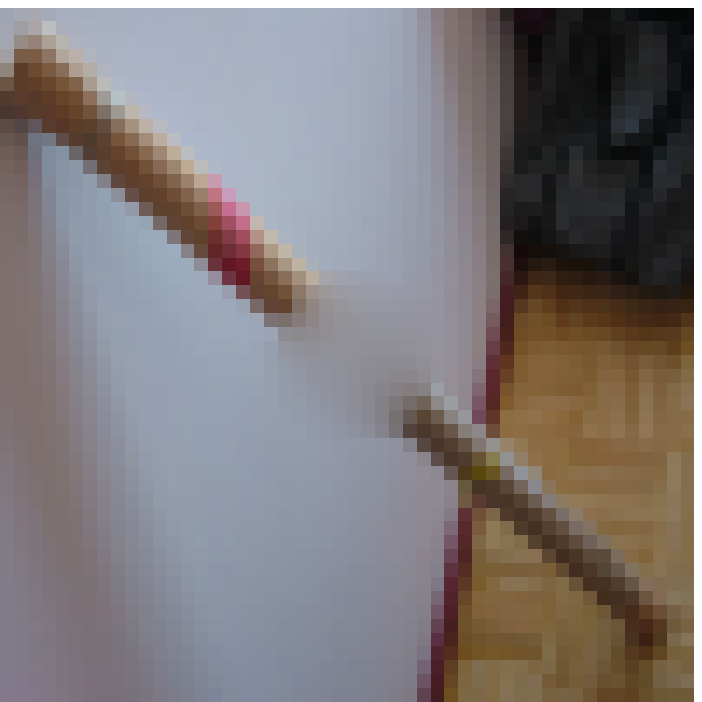} &
\includegraphics[width = \linewidth]{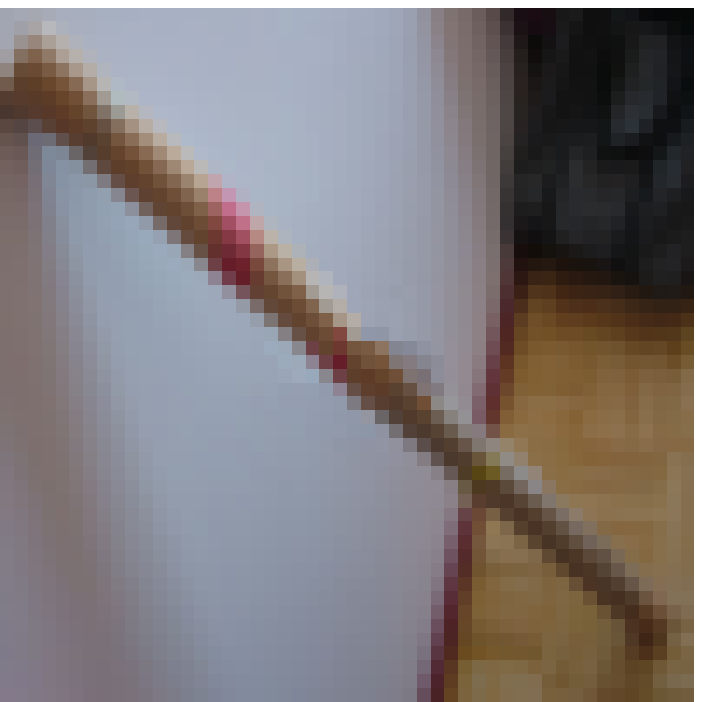} \\
\begin{center}(a) Occluded input image\end{center} &
\begin{center}(b) Random initialisation\end{center} &
\begin{center}(c) Smooth zero-Laplacian interpolation\end{center} &
\begin{center}(d) Onion peel initialisation\end{center} \\
\includegraphics[width = \linewidth]{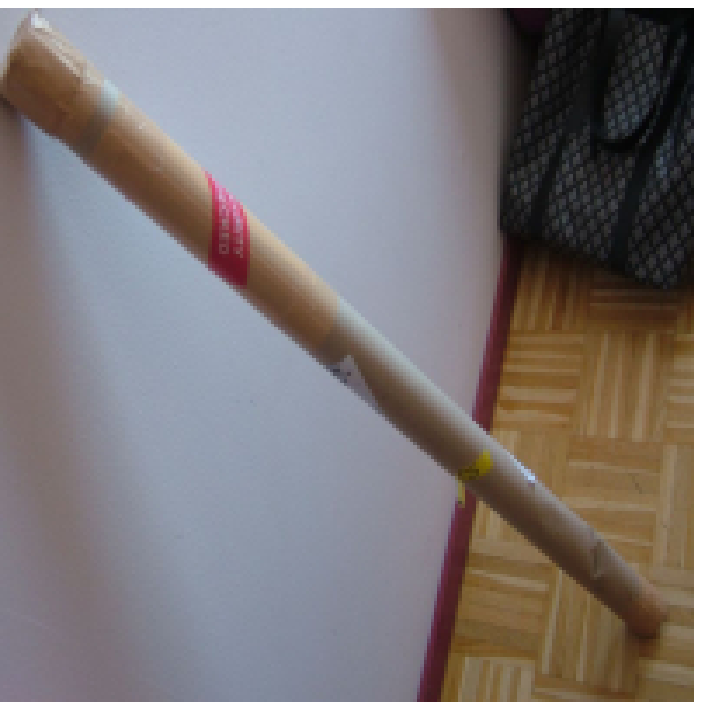} &
\includegraphics[width = \linewidth]{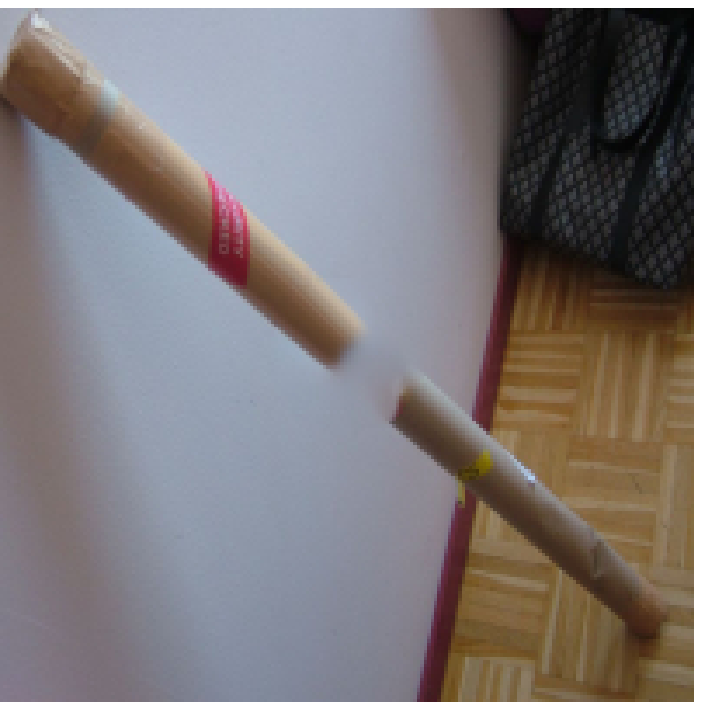} &
\includegraphics[width = \linewidth]{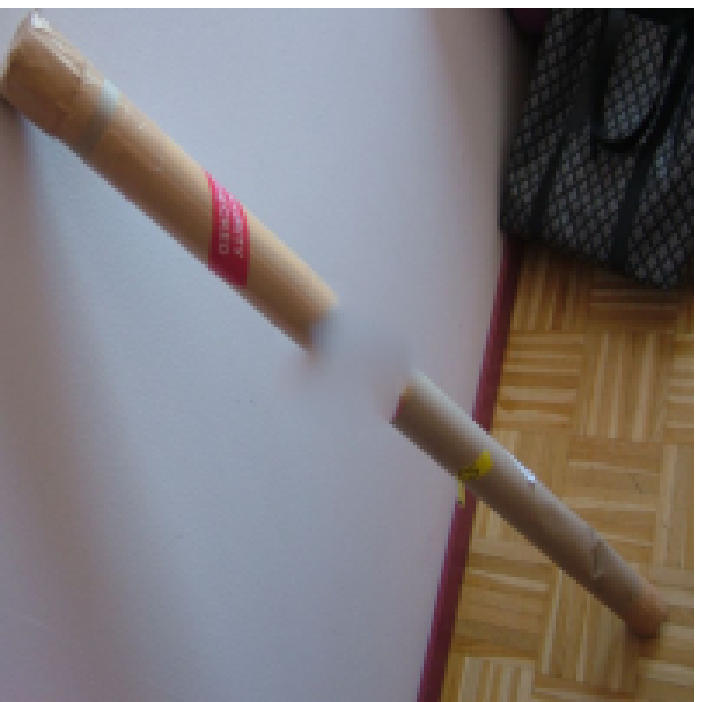}  &
\includegraphics[width = \linewidth]{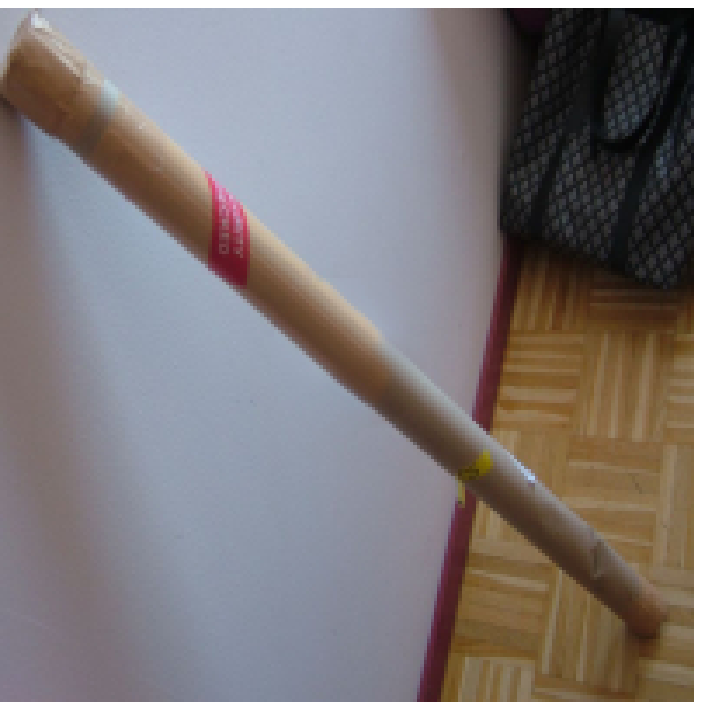} \\
\begin{center}(a) Original image\end{center} &
\begin{center}(b) Inpainting after random initialisation\end{center} &
\begin{center}(c) Inpainting after zero Laplacian\end{center} &
\begin{center}(d) Inpainting after onion peel initialisation\end{center} \\
\end{tabularx}
\caption{{\bf Impact of initialisation schemes on inpainting results}.
		Note that the only scheme which successfully reconstructs the cardboard tube is the ``onion peel'' approach where first filling at the coarsest resolution is conducted in a greedy fashion.}
\label{fig:onionPeelInitialisation}
\end{figure}


The iterative procedure at the heart of our algorithm relies
on an initial inpainting solution.
The initialisation step is very often left unspecified in work on
video inpainting. As we shall see in this Section, it plays a vital role,
and we therefore explain our chosen initialisation method in detail.

We inpaint
at the coarsest level using an ``onion peel'' approach,
that is to say we inpaint one layer of the occlusion at a time,
each layer being one pixel thick.

More formally, let $\cH'\subset\cH$ be the current occlusion,
and $\partial \cH'\subset\cH'$
the current layer to inpaint.
We define the \emph{unoccluded} neighbourhood $\N'_p$ of a pixel
$p$, with respect to the current occlusion $\cH'$ as:
\begin{equation}
\N_p' = \{q \in \N_p, q \notin \cH'\}.
\end{equation}

Some choices are needed to implement this initialisation method. First
of all, we only compare the unoccluded pixels during a patch
comparison. The distance between two patches $W_p$ and $W_{p+\phi(p)}$
is therefore redefined as:
\begin{equation}
d^2(W_p,W_{p+\phi(p)}) =  \frac{1}{|\N'_p|} \sum_{q \in \N'_p} \Big(\|u(q) - u(q+\phi(p))\|_2^2 + \lambda \| T(q) - T(q+\phi(p)) \|_2^2 \Big).	
\label{eq:patchDistancePartial}					
\end{equation}

We also need to choose which neighbouring patches to use
for reconstruction. Some will be quite unreliable, as
only a small part of the patches are compared.
In our implementation, we only use the ANNs of patches whose centres are located
outside the current occlusion layer.
Formally, we reconstruct the pixels in the current layer
by using the following formula, modified from Equation~\ref{eq:wexlerWeightedMean}:

\begin{equation}
u_{p} =
	\frac{ \sum_{q \in \N'_p} s_p^q u(p + \phi(q))}
			{\sum_{q \in \N'_p} s_p^q}.
\label{eq:initialisationWexler}
\end{equation}
The same reconstruction is applied to the texture features.
A pseudo-code for the initialisation procedure may be seen in Algorithm~\ref{algo:inpaintingInitialisation}.

\begin{algorithm}
\SetAlgoLined
\KwData{Coarse level inputs $u^L$, $T^L$, $\phi^L$, $\cH^L$}
 \KwResult{Coarse initial filling $u^L$, $T^L$ and map $\phi^L$}
 \BlankLine
 $B \leftarrow 3 \times 3 \times 3$ structuring element\;
 $\cH' \leftarrow \cH^L$\;
 \While{$\cH' \neq \emptyset$}
 {
 	$\partial \cH' \leftarrow$ $\cH'\setminus$Erosion($\cH', B$)\;
	$\phi^L \leftarrow$ ANNsearch($u^L,T^L,\phi^L,\partial \cH'$) \tcp*[r]{Alg.\ref{algo:ANNsearch}, with partial distance (\ref{eq:patchDistancePartial})}
	$u^L \leftarrow$ Reconstruction($u^L,\phi^L,\partial \cH'$) \tcp*[r]{Eq.\ref{eq:initialisationWexler}}
	$T^L \leftarrow$ Reconstruction($T^L,\phi^L,\partial \cH'$) \tcp*[r]{Eq.\ref{eq:initialisationWexler}}
 	$\cH' \leftarrow$ Erosion($\cH', B$)\;
 }
 \caption{Inpainting initialisation}
 \label{algo:inpaintingInitialisation}
\end{algorithm}

Figure~\ref{fig:onionPeelInitialisation} shows some evidence to support
a careful initialisation scheme. Three different initialisations have been
tested: random initialisation, zero-Laplacian interpolation and onion
peel. Random initialisation is achieved by initialising the occlusion
with pixels chosen randomly from the image. Zero Laplacian (harmonic) interpolation
is the solution of the Laplace equation $\Delta u = 0$ with Dirichlet
boundary conditions stemming from $\D$.
It may be seen (Figure~\ref{fig:onionPeelInitialisation}) that the
first two initialisations are unable to join the two parts of the cardboard tube
together, and
that the subsequent iterations do not improve the situation.
In contrast, the proposed initialisation produces
a satisfactory result.

In order to make our method as easy as possible to reimplement,
we now present some further algorithmic details, which are
in fact very important for achieving good results.

\subsection{Implementation of the multi-resolution scheme}
\label{subsec:multiResolutionDetails}

Our first remarks concern the implementation of the multi-resolution
pyramids. Wexler \emph{et al}. and Granados \emph{et al}. both note that \emph{temporal} subsampling
can be detrimental to inpainting results. This is due to the difficulty of
representing motion at coarser levels. For this reason
we do not subsample in the temporal direction, as in \cite{Granados2012How}.
The only case where we need to temporally subsample is when the objects spend
a long time behind the occlusion, (this was only done in the ``Jumping girl'' sequence).
This is quite a hard problem to solve since it becomes increasingly
difficult to decide what motion an occluded object should have when
the occlusion time grows longer, unless there is strictly periodic
motion. We leave this as an open question, which could be investigated in further work.

A crucial choice when using multi-resolution schemes is the number of
pyramid levels to use. Most other methods leave this parameter unspecified,
or fix the size of the image/video at the coarsest scale, and
determine the resulting number of levels \cite{Granados2012How,Liu2013ExemplarBased,Pritch2009Shiftmap}.
In fact, when one considers the problem in more detail, it becomes apparent
that the number of levels should be set so that the occlusion
size is not too large in comparison to the patch size. This
intuition is supported by experiments in
very simple image inpainting situations, which showed that the occlusion
size should be somewhat less than \emph{twice} the patch size.
In our experiments, we follow this general rule of thumb.

\begin{figure}[!ht]
\begin{tabular}{c c c}
\includegraphics[width = 0.3\linewidth]{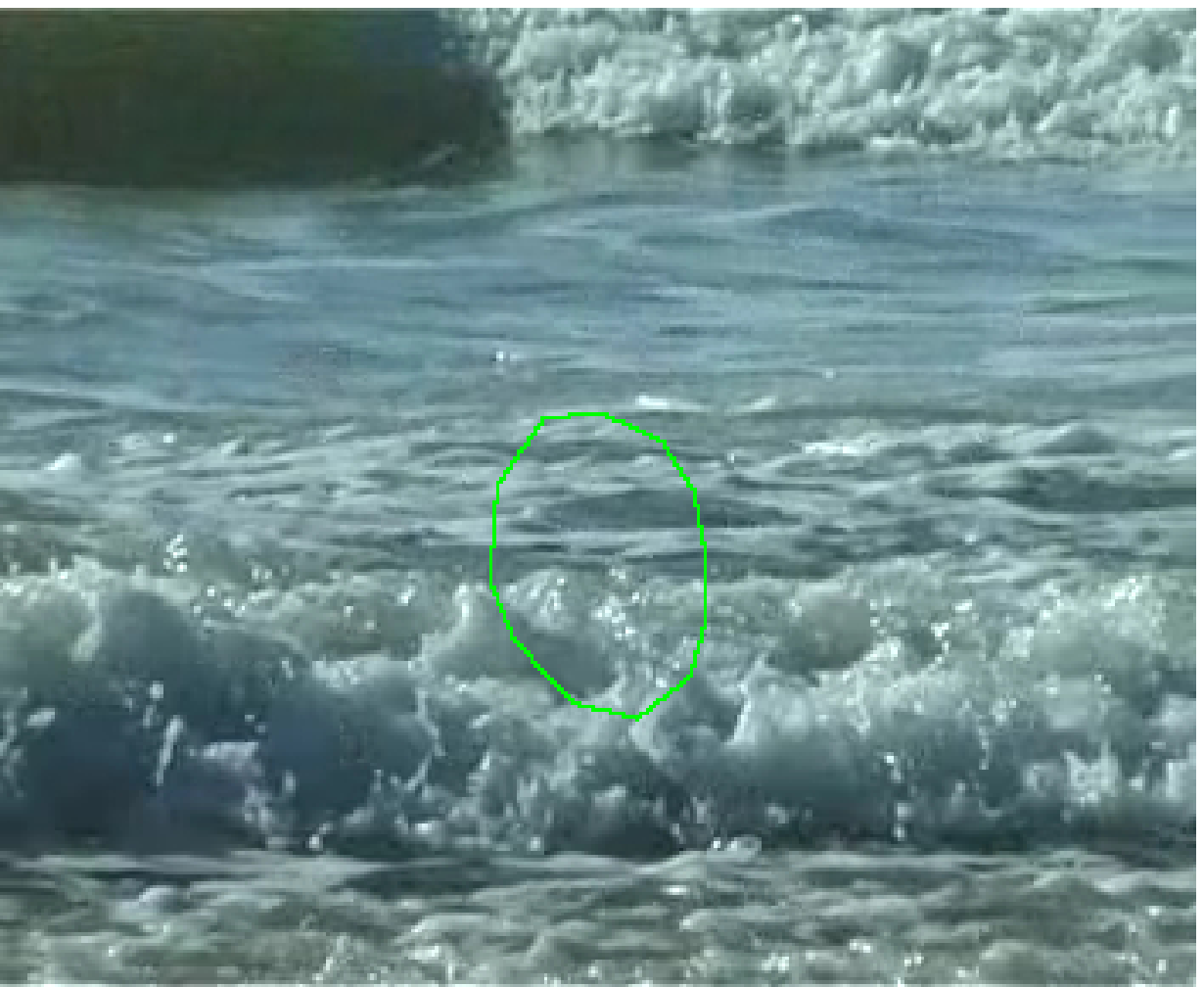} &
\includegraphics[width = 0.3\linewidth]{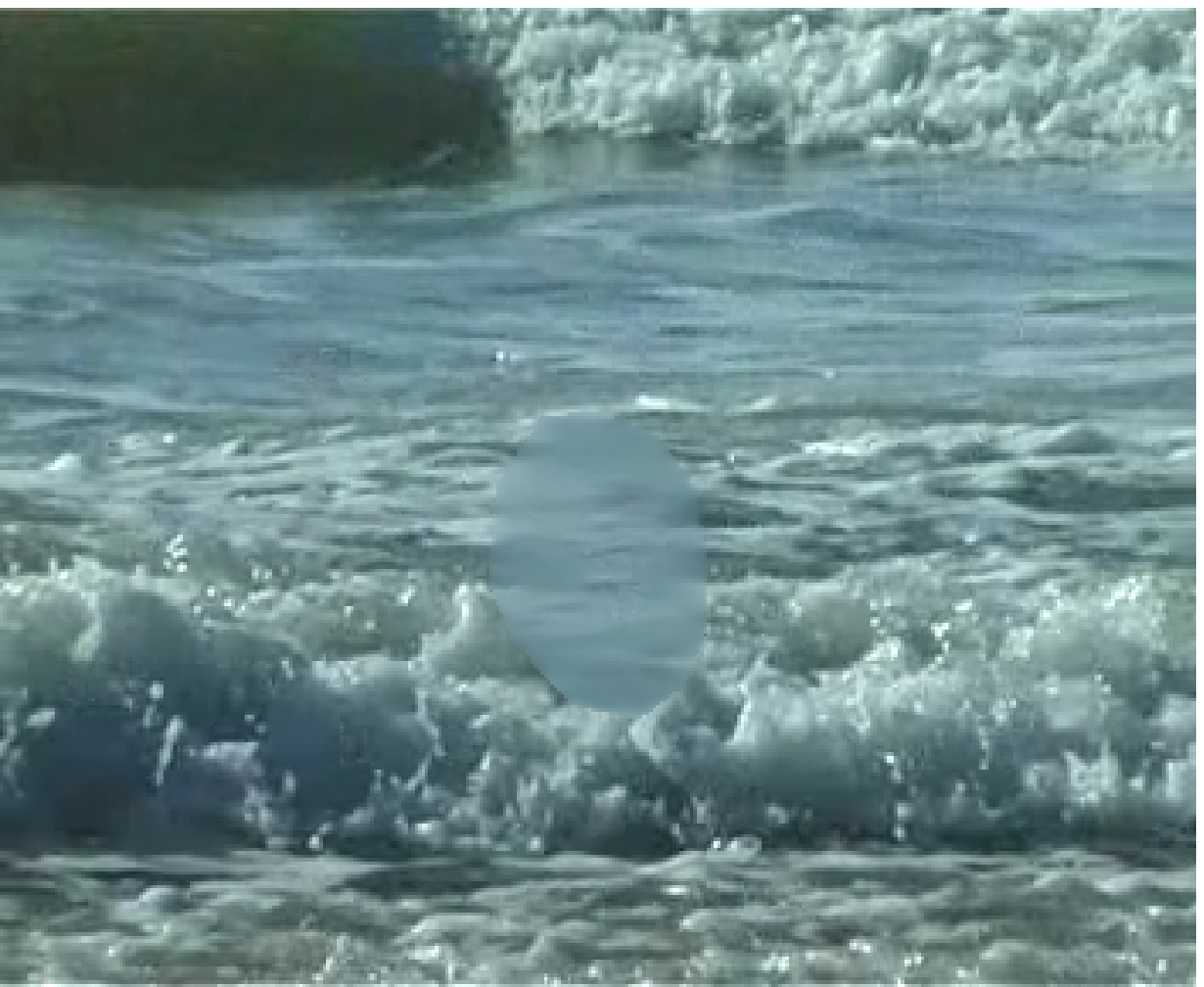} &
\includegraphics[width = 0.3\linewidth]{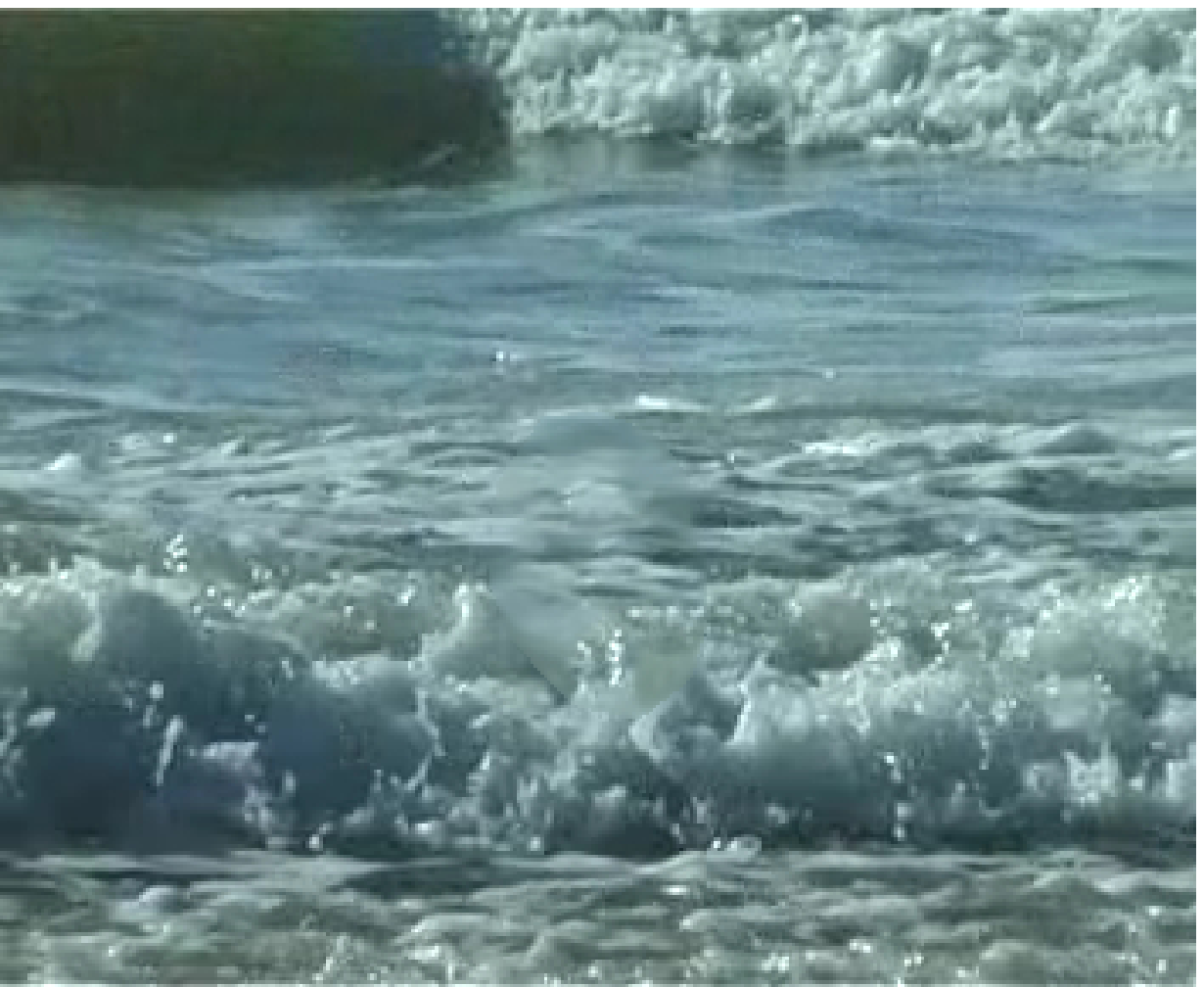} \\
\centering{Original frame : ``Waves''} & Inpainting without features & Inpainting with features\\
\end{tabular}
\caption{{\bf Usefulness of the proposed texture features}. Without the
features, the algorithm fails to recreate correctly the waves, which is a typical example of complex video texture.} 
\label{fig:videoTexture}
\end{figure}


Another question which is of interest is how to pass from one pyramid
level to another. Wexler \emph{et al}. presented quite an intricate scheme
in \cite{Wexler2007SpaceTime} to do this, whereas Granados \emph{et al}. propose
a simple upsampling of the shift map. This is conceptually simpler than the approach of Wexler \emph{et al}.
and, after experimentation, we chose this option as well. Therefore,
the shift map $\phi$ is upsampled using nearest neighbours interpolation,
and both the higher resolution video and the higher resolution texture
features are reconstructed using Equation~\ref{eq:wexlerWeightedMean}.
One final note on this point is that
we use the upsampled version of $\phi$ as an initialisation for the PatchMatch
algorithm at each level apart from the coarsest (this differs to our
previous work in \cite{Newson2013Towards}).

\subsection{Various implementation details}

We also require a threshold which will stop the iterations
of the ANN search and reconstruction steps.
In our work, we use the average
colour difference in each channel per pixel between iterations
as a stopping criterion.
If this falls below a certain threshold, we stop
the iteration at the current level. We set this threshold
to 0.1. In order to avoid iterating for too long,
so we also impose a maximum number of 20 iterations at any
given pyramid level.

The patch size parameters were set to $5 \times 5 \times 5$ 
in all of our experiments.
We set the texture feature parameter $\lambda$ to 50.
Concerning the spatio-temporal PatchMatch, we use ten iterations of
propagation/random search during the ANN search algorithm
and set the window size reduction factor $\beta$
to 0.5 (as in the paper of Barnes \emph{et al}. \cite{Barnes2009PatchMatch}).

The complete algorithm is summarized in Alg. \ref{algo:completeAlgo}.

\begin{figure}[!ht]
\centering
\begin{tabular}{c c c}
\includegraphics[width = 0.3\linewidth]{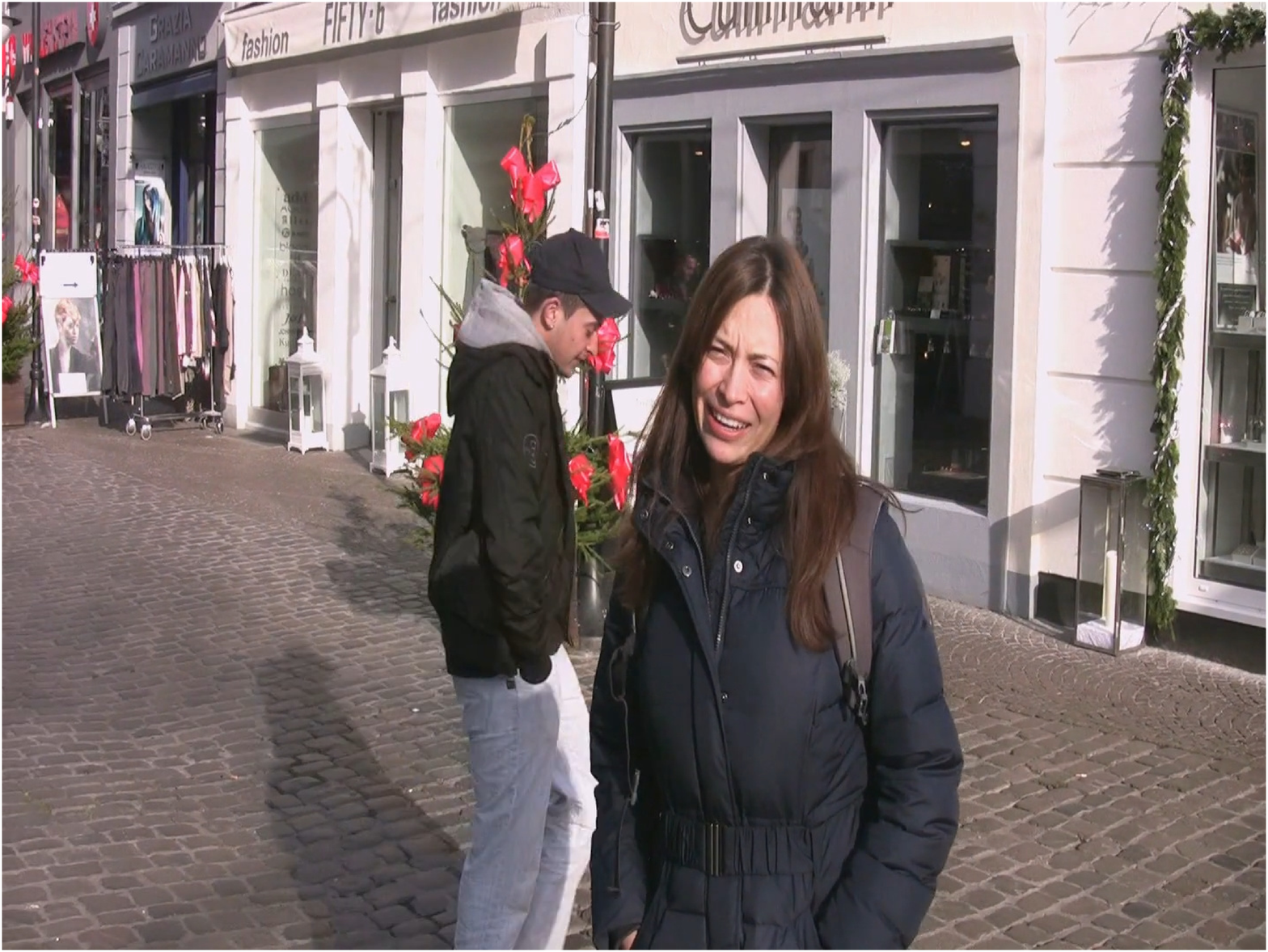} &
\includegraphics[width = 0.3\linewidth]{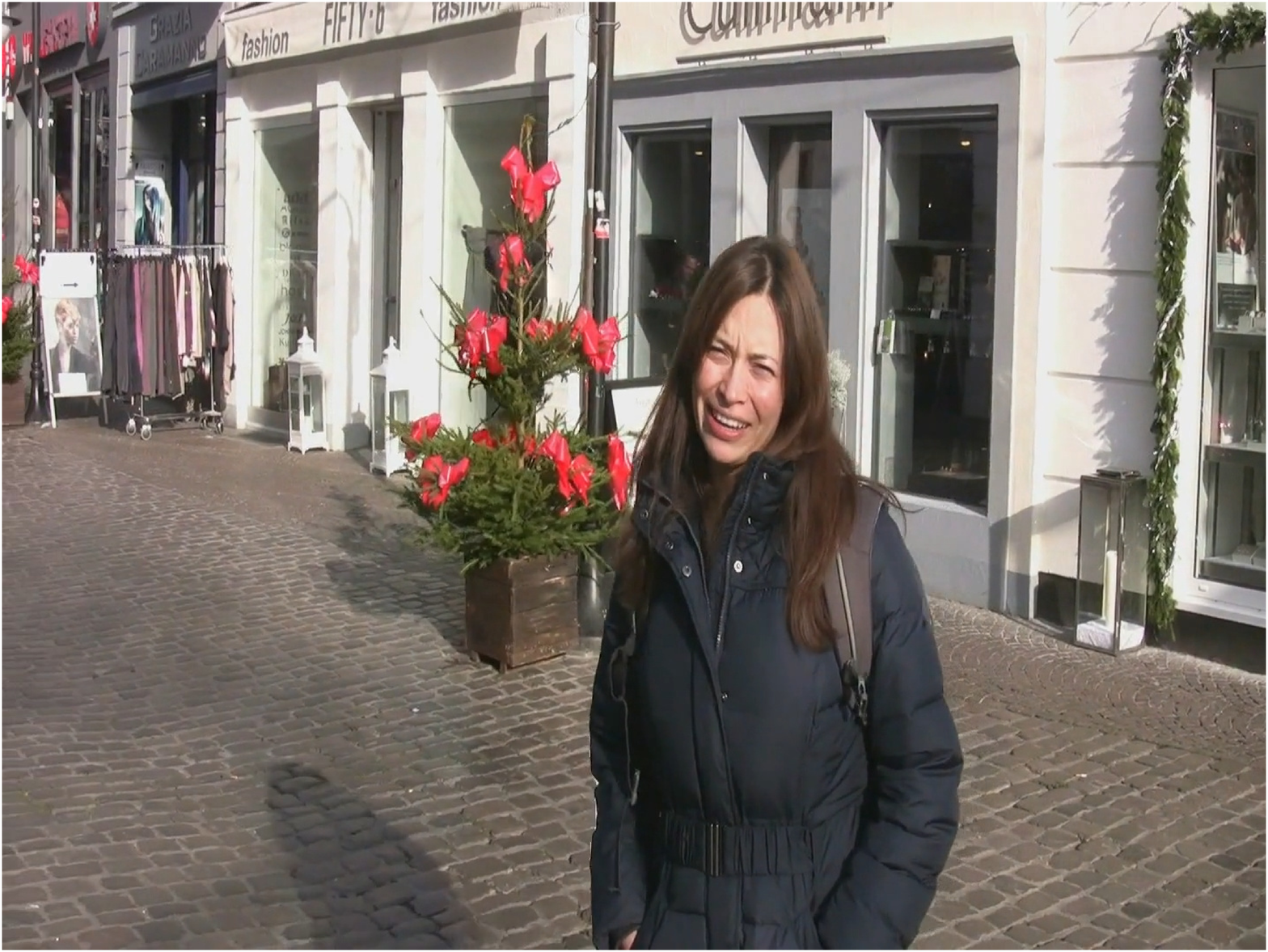} &
\includegraphics[width = 0.3\linewidth]{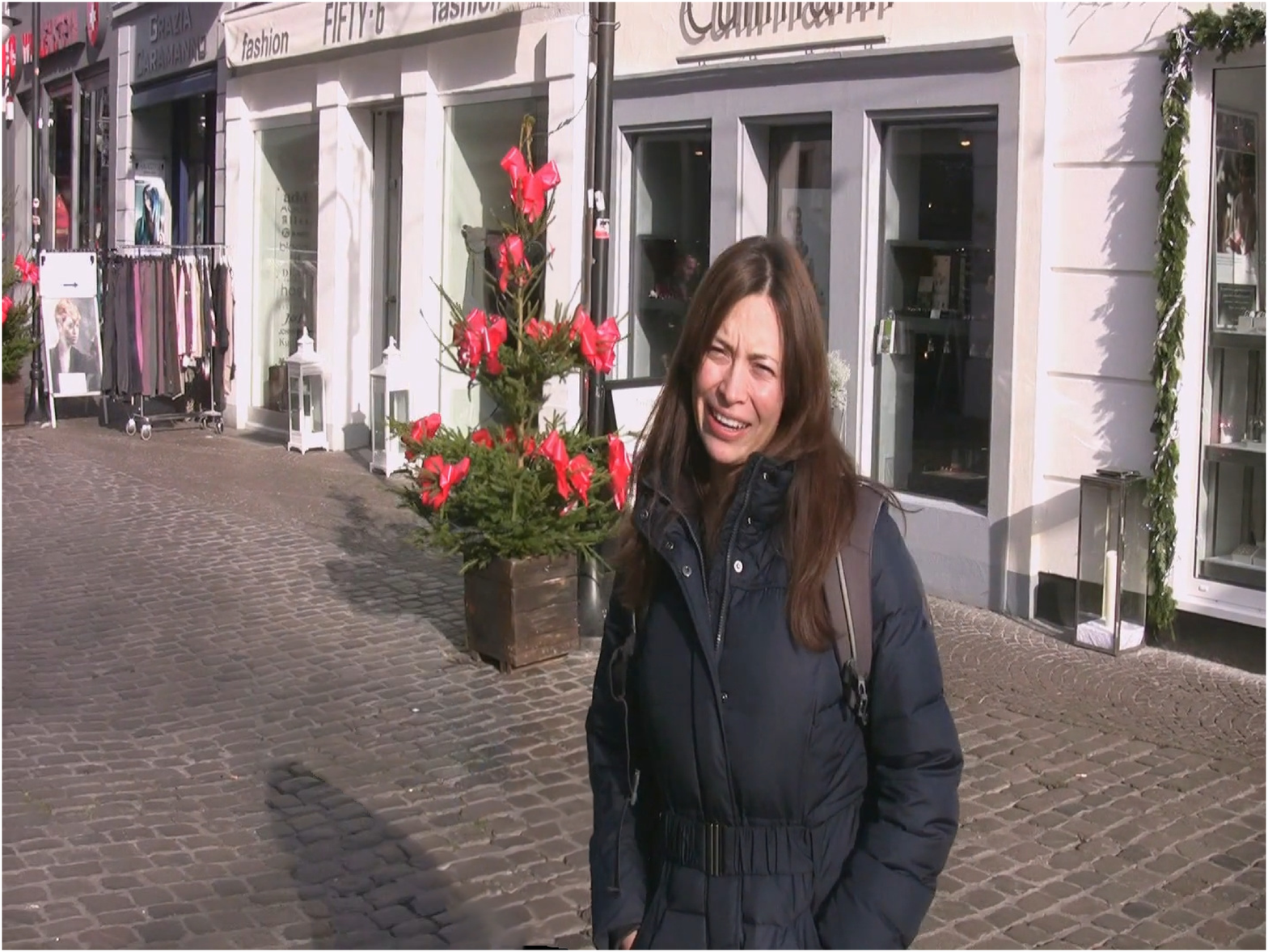} \\
\centering{Original frame : ``Girl'' \cite{Granados2012Background}} & Result from \cite{Granados2012Background} & Our result\\
\end{tabular}
\caption{{\bf A comparison of our inpainting result with the that of the background inpainting algorithm of
		Granados \emph{et al.} \cite{Granados2012Background}}. In such cases with moving background,
		we are able to achieve high quality results (as do Granados \emph{et al}.), but we
		do this in one, unified algorithm. This illustrates the capacity of our algorithm
		to perform well in a wide range of inpainting situations.
		}
\label{fig:realignmentComparison}
\end{figure}

\begin{algorithm}
\SetAlgoLined
\KwData{Input video $u$ over $\Omega$, occlusion $\cH$, resolution number $L$}
\KwResult{Inpainted video}
\BlankLine
$(u,\theta)\leftarrow$AlignVideo($u$) \tcp*[r]{Alg.\ref{algo:realignmentPreProcessing}}
$\{u^{\ell}\}_{\ell=1}^L \leftarrow$ImagePyramid($u$)\;
$\{T^{\ell}\}_{\ell=1}^L \leftarrow$TextureFeaturePyramid($u$)\tcp*[r]{Eqs.\ref{eq:textureFeature}-\ref{eq:texturePyr}}
$\{\cH^{\ell}\}_{\ell=1}^L \leftarrow$OcclusionPyramid($\cH$)\;
$\phi^L\leftarrow$Random\;
$(u^L,T^L,\phi^L)\leftarrow$Initialisation($u^L,T^L,\phi^L,\cH^L$)\tcp*[r]{Alg.\ref{algo:inpaintingInitialisation}}
\For{$\ell = L$ \KwTo $1$} 
{
	$k=0$, $e=1$\;
	\While{$e> 0.1$ {\bf and} $k< 20$}
	{
			$v=u^{\ell}$\;					
			$\phi^{\ell} \leftarrow $ANNsearch$(u^{\ell},T^{\ell},\phi^{\ell},\cH^{\ell})$ \tcp*[r]{Alg.\ref{algo:ANNsearch}}
			$u^{\ell} \leftarrow $Reconstruction$(u^{\ell},\phi^{\ell},\cH^{\ell})$ \tcp*[r]{Eqs.\ref{eq:wexlerWeightedMean}-\ref{eq:wexlerWeights}}
			$T^{\ell} \leftarrow $Reconstruction$(T^{\ell},\phi^{\ell},\cH^{\ell})$\;
			$e=\frac{1}{3|\cH^{\ell}|}\|u_{\cH^{\ell}}^{\ell}-v_{\cH^{\ell}}\|_2$\;			
			$k \leftarrow k+1$\;
	}
	\eIf{$\ell=1$}
	{
	$u \leftarrow $FinalReconstruction$(u^{1},\phi^{1},\cH)$ \tcp*[r]{Eq.\ref{eq:finalReconstruction}}
	}
	{
	$\phi^{\ell -1} \leftarrow $UpSample$(\phi^{\ell},2)$ \tcp*[r]{Sec.\ref{subsec:multiResolutionDetails}}
	$u^{\ell -1} \leftarrow $Reconstruction$(u^{\ell -1},\phi^{\ell -1},\cH^{\ell -1})$\;
	$T^{\ell -1} \leftarrow $Reconstruction$(T^{\ell -1},\phi^{\ell -1},\cH^{\ell -1})$\;
	} %
}
$u \leftarrow$UnwarpVideo($u,\theta$)
\caption{Proposed video inpainting algorithm.}
\label{algo:completeAlgo}
\end{algorithm}


\section{Experimental results}
\label{sec:results}
The goal of our work is to achieve high quality inpainting results in
varied, complex video inpainting situations, with reduced execution time.
Therefore, we shall evaluate our results in terms of visual quality and
execution time.

We compare our
work to that of Wexler \emph{et al.} \cite{Wexler2007SpaceTime}
and to the most recent video inpainting
method of Granados \emph{et al}. \cite{Granados2012How}.
All of the videos in this paper (and more) can be viewed and downloaded along with occlusion masks at 
\url{http://www.telecom-paristech.fr/~gousseau/video_inpainting}. An implementation of our method is also
available at this address.

\subsection{Visual evaluations}


\begin{figure}
\begin{tabularx}{\linewidth}{X X X}
\multicolumn{3}{c}{``Beach Umbrella''} \\ 
\includegraphics[width = \linewidth, height=20mm]{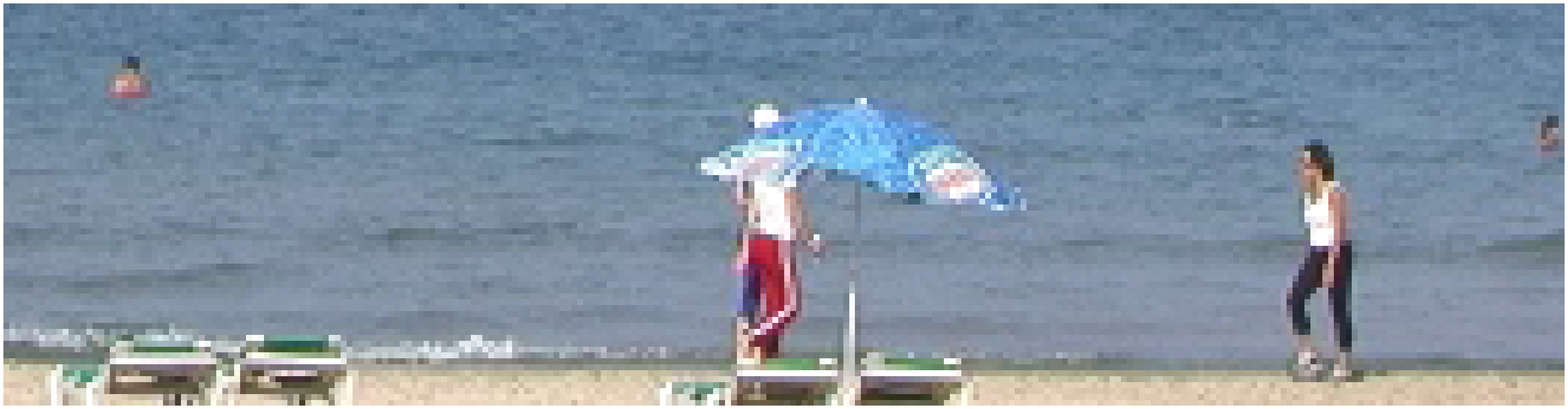} &
\includegraphics[width = \linewidth, height=20mm]{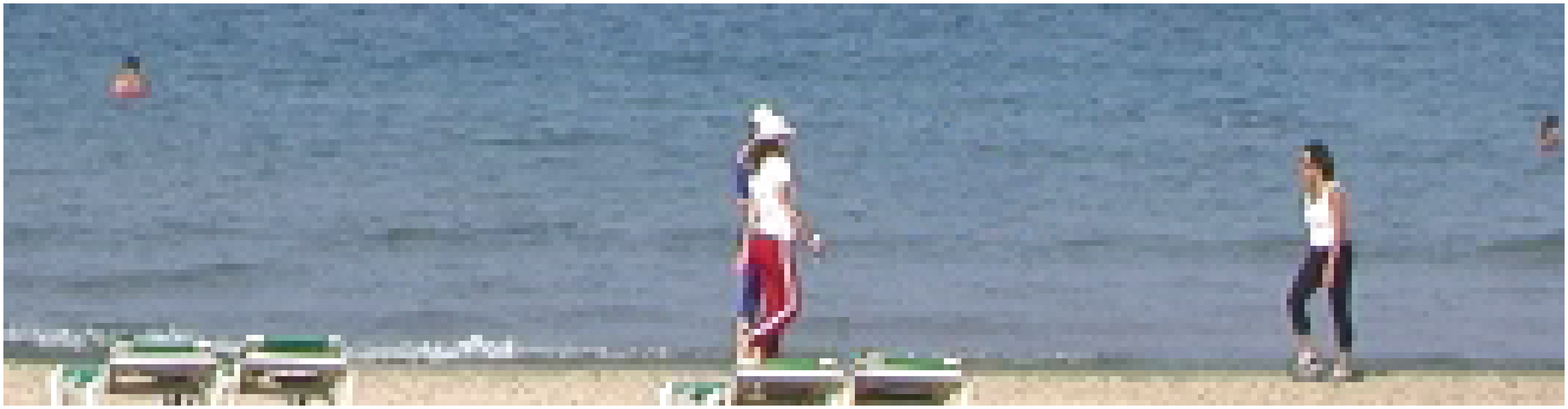} &
\includegraphics[width = \linewidth, height=20mm]{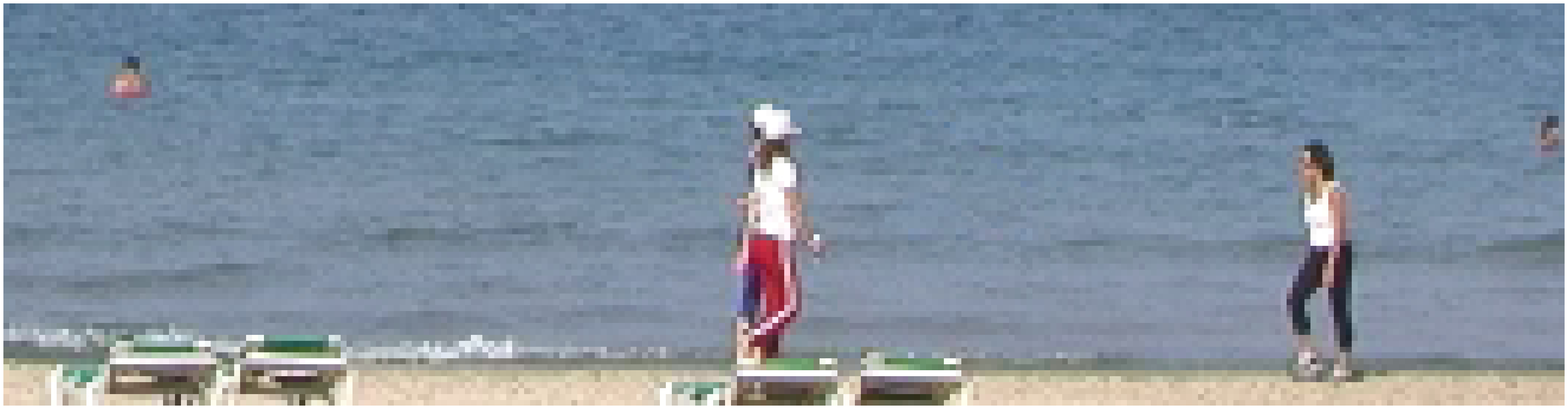} \\
& & \\
\multicolumn{3}{c}{``Crossing Ladies''} \\
\includegraphics[width = \linewidth, height=25mm]{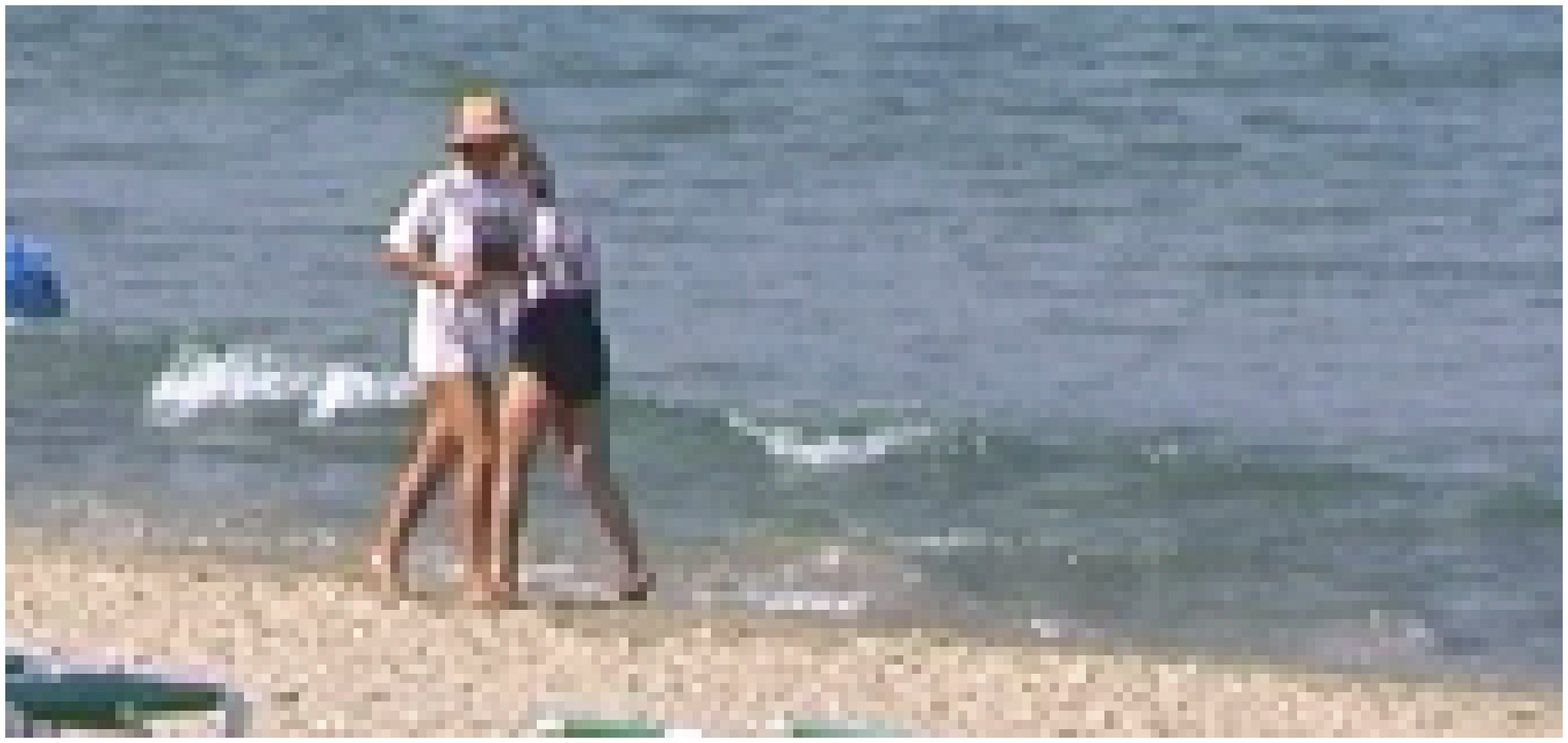} &
\includegraphics[width = \linewidth, height=25mm]{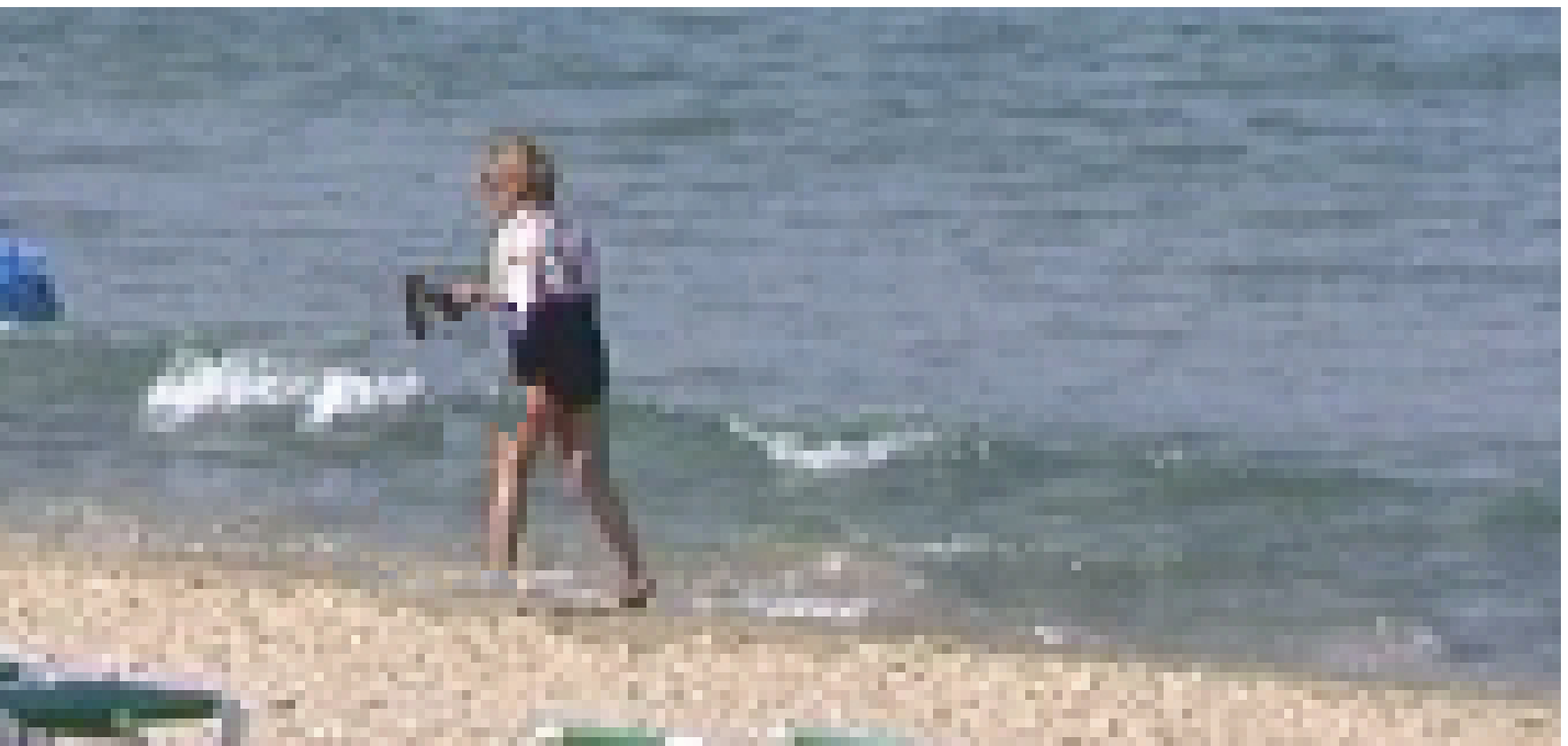} &
\includegraphics[width = \linewidth, height=25mm]{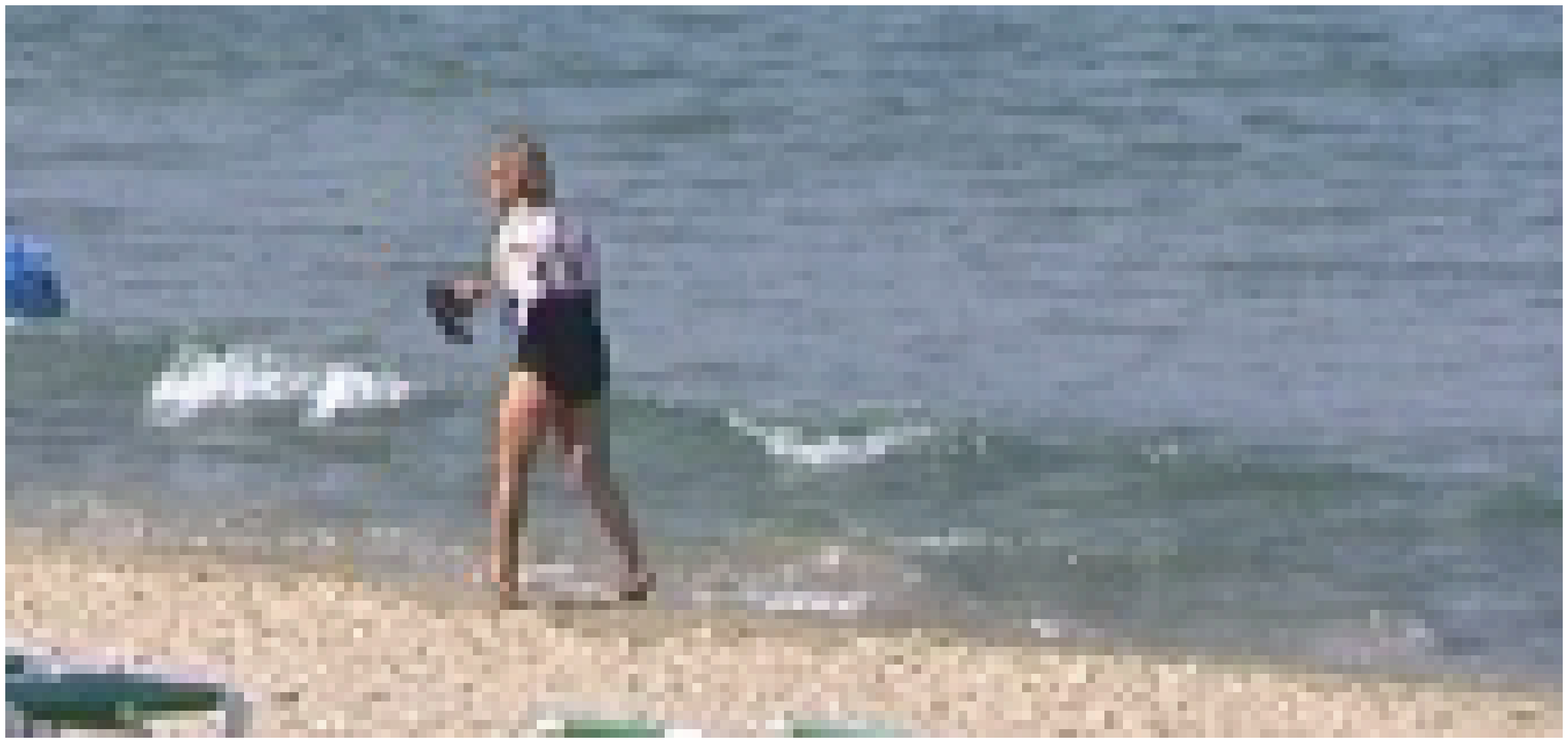} \\
& & \\
\multicolumn{3}{c}{``Jumping girl''} \\
\includegraphics[width = \linewidth, height=20mm]{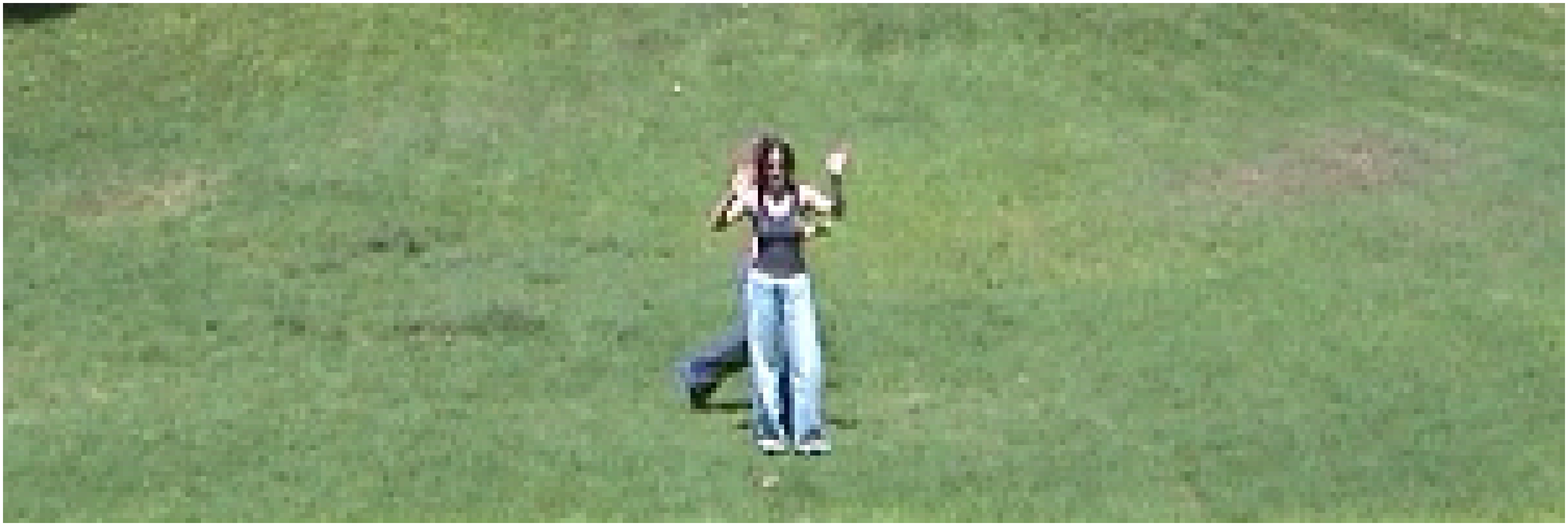} &
\includegraphics[width = \linewidth, height=20mm]{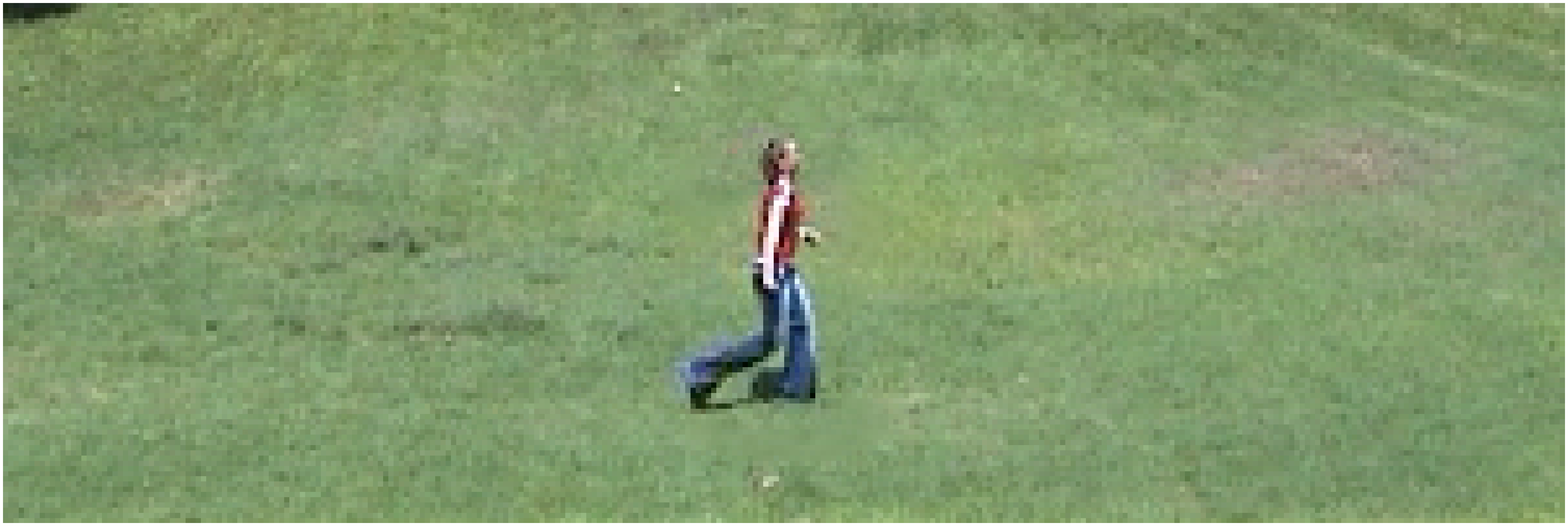} &
\includegraphics[width = \linewidth, height=20mm]{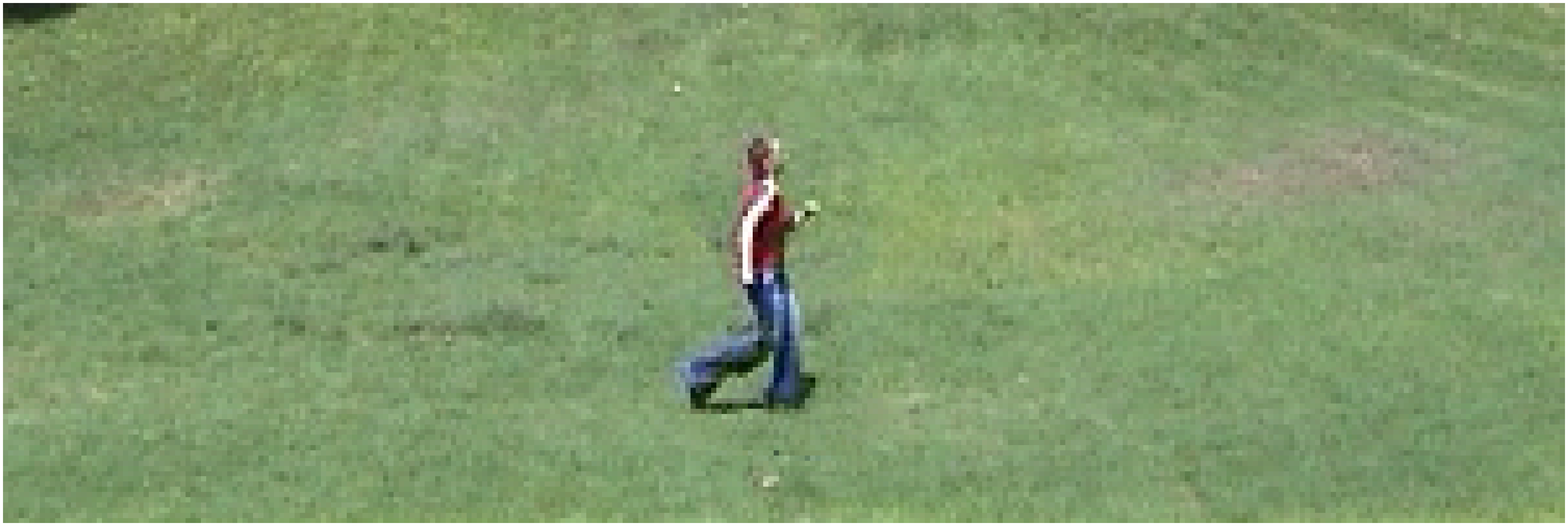} \\
\centering{Original frames}
& \centering{Inpainting result from \cite{Wexler2007SpaceTime}} & \centering{Our inpainting result}
\end{tabularx}
\caption{{\bf Comparison with Wexler \etal} We achieve
results of similar visual quality compared to those in \cite{Wexler2007SpaceTime}, with a reduction of the ANN search time by a factor of up to 50 times.}
\label{fig:wexlerInpaintingResults}
\end{figure}



\begin{figure}
\centering
\begin{tabular}{c c}
\centering{Original frame : ``Duo''} & Original frame : ``Museum'' \\
\includegraphics[width = 0.45\linewidth]{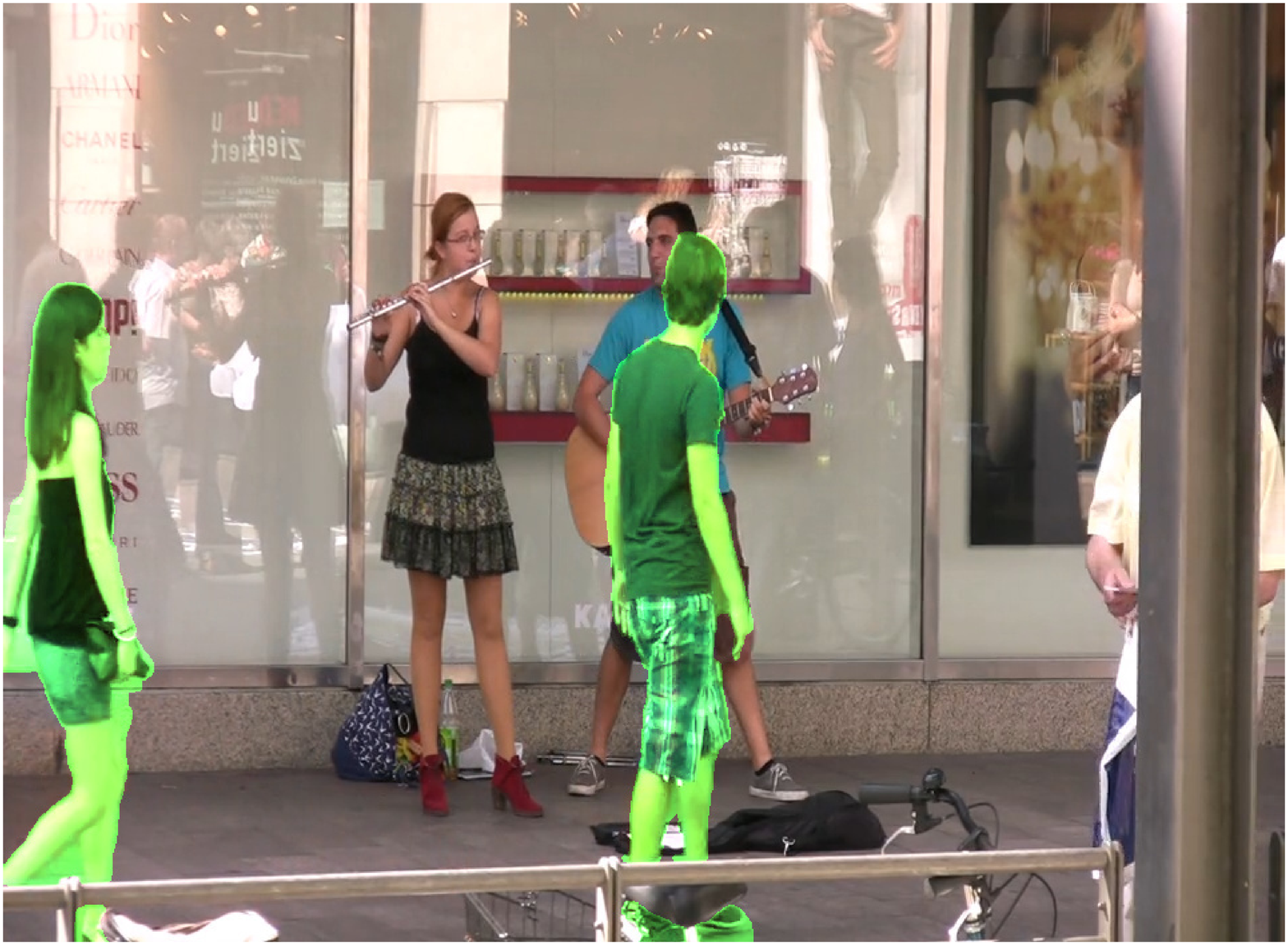} &
\includegraphics[height = 51mm]{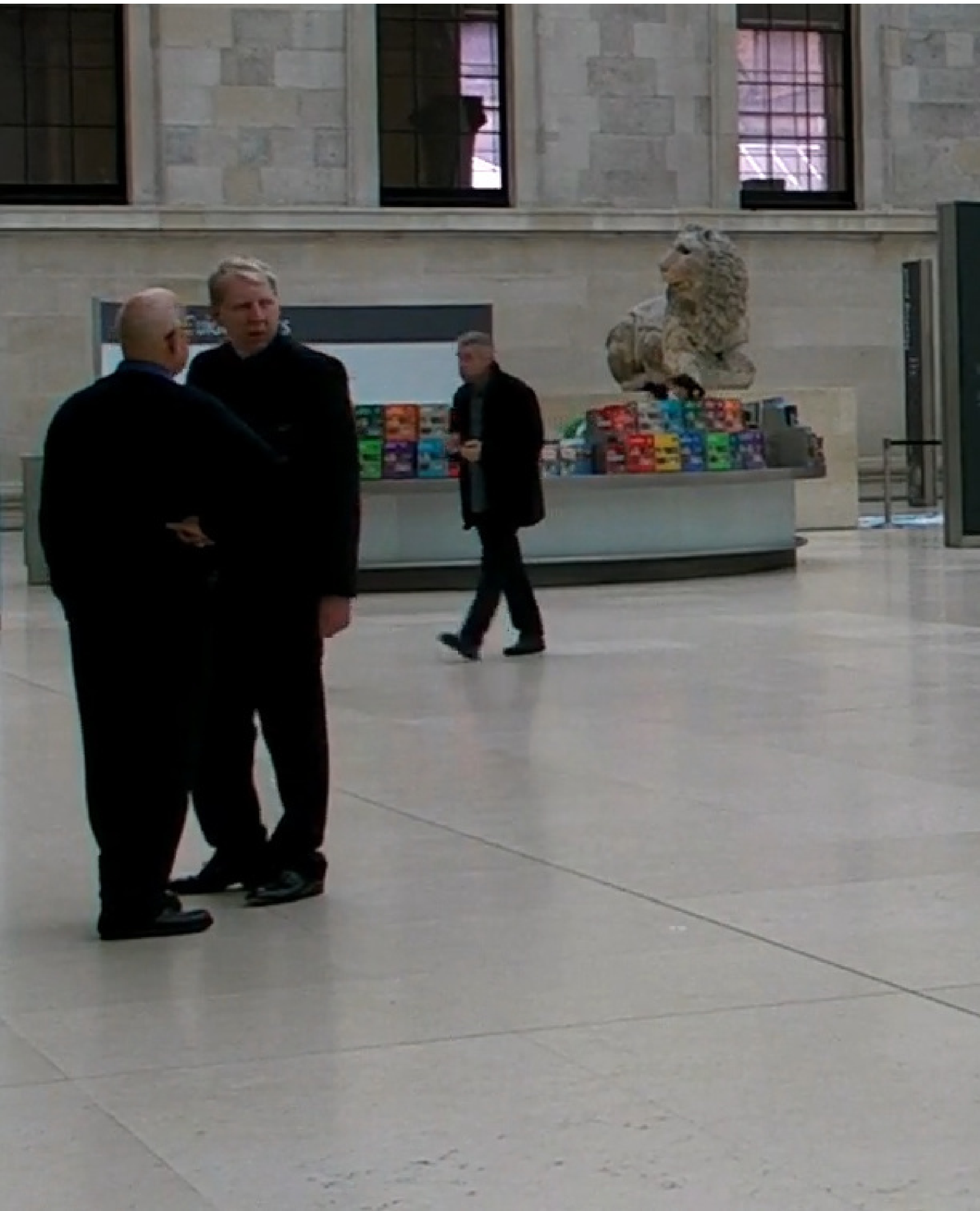} \\
&  \\ &  \\&  \\&  \\
 \multicolumn{2}{c}{Inpainting result from \cite{Granados2012How}} \\
\includegraphics[width = 0.45\linewidth]{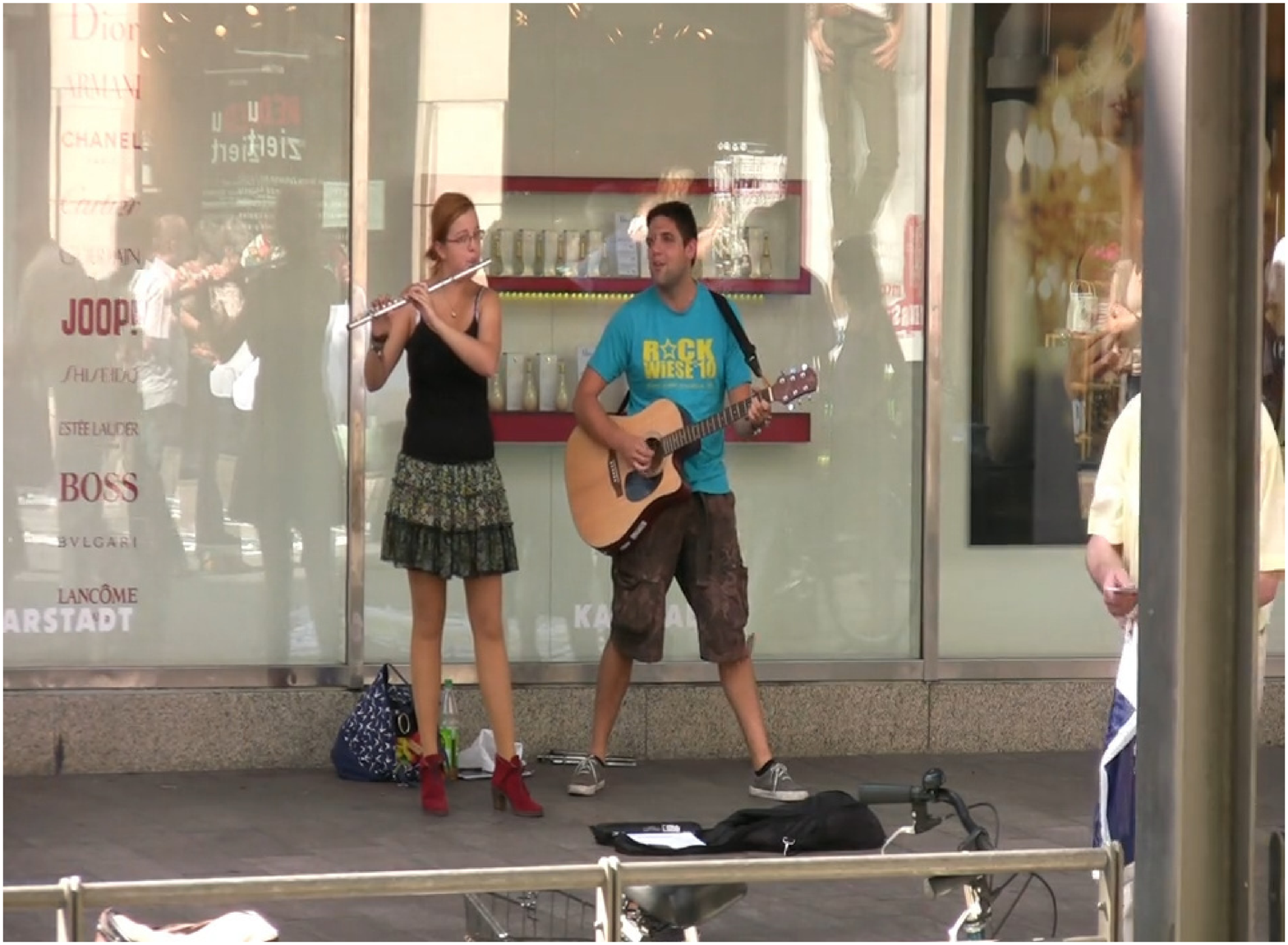} &
\includegraphics[height = 51mm]{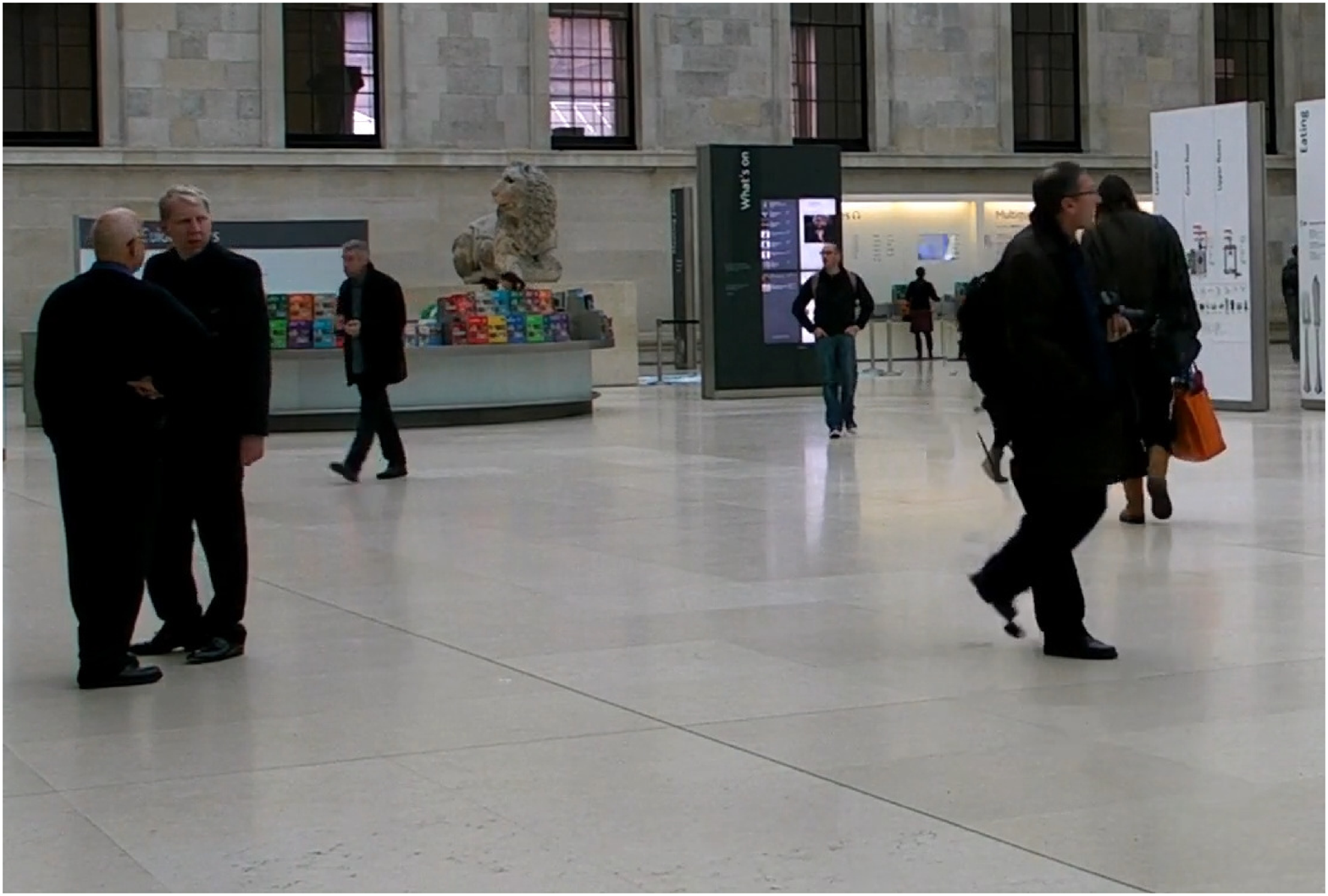} \\
&  \\ & \\
\multicolumn{2}{c}{Our inpainting result} \\
\includegraphics[width = 0.45\linewidth]{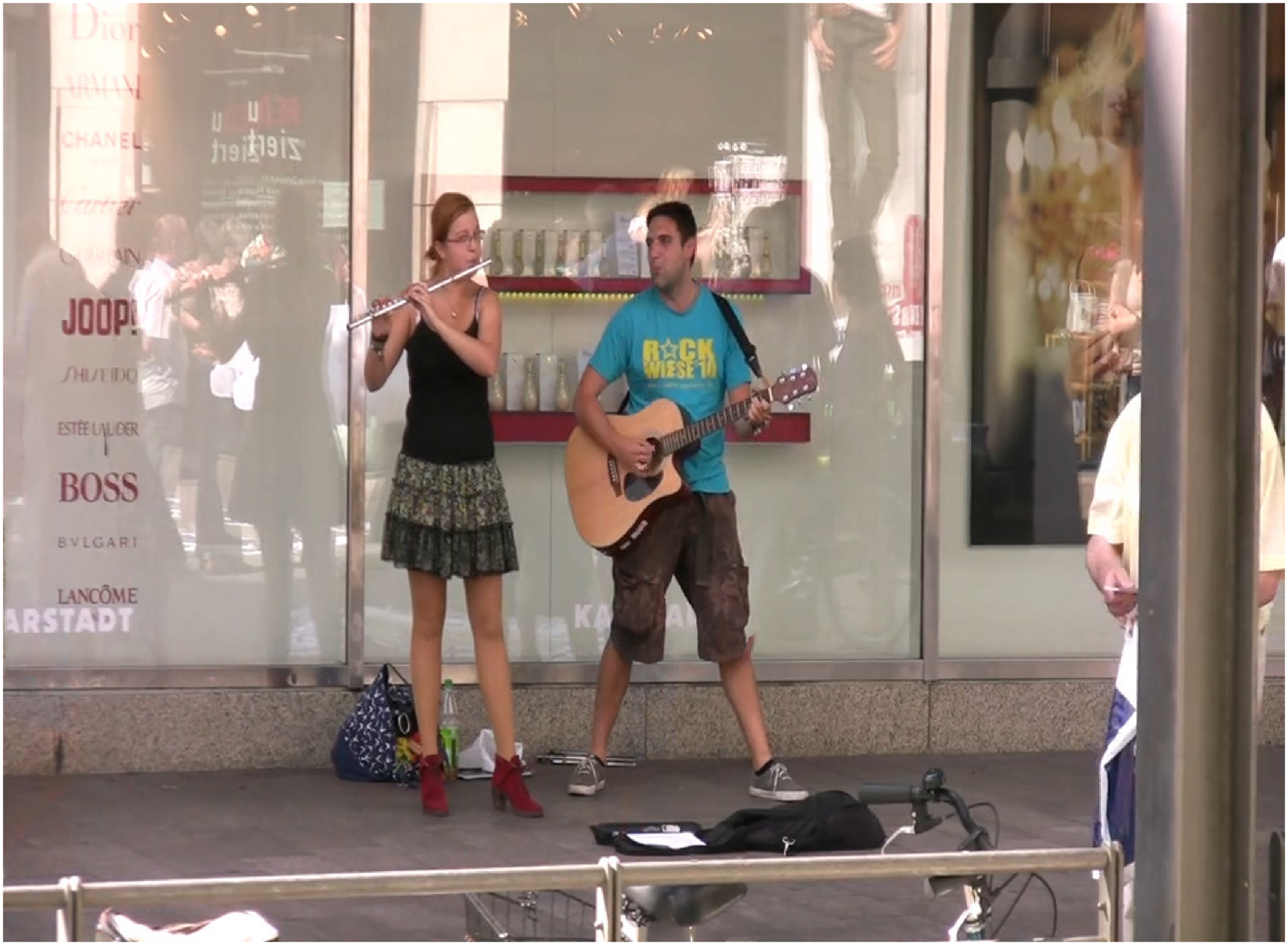} &
\includegraphics[ height = 51mm]{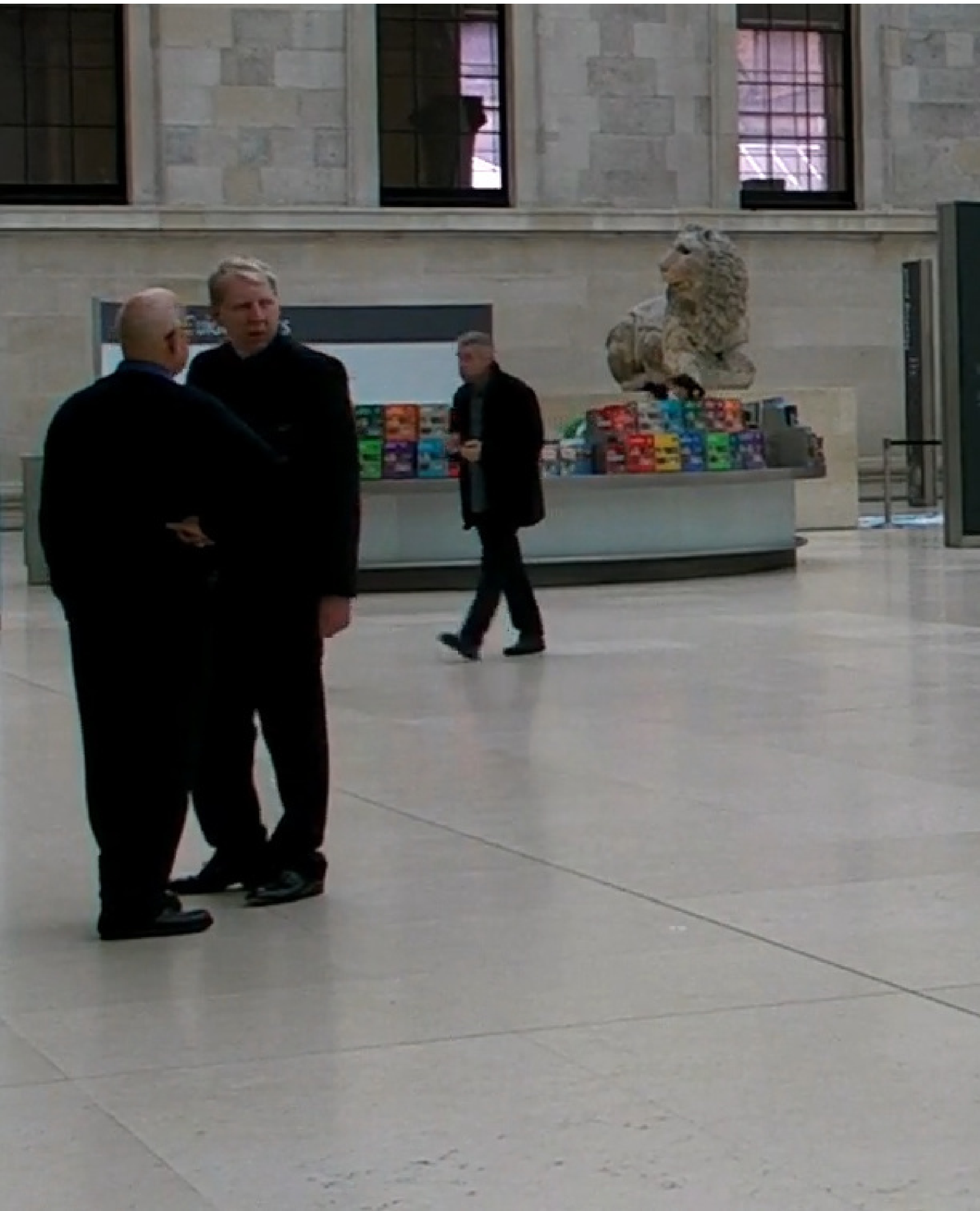} \\
\end{tabular}
\caption{{\bf Comparison with Granados \etal} We achieve similar results to those of \cite{Granados2012How} in
an order of magnitude less time, without user intervention. The occlusion masks are highlighted in green.}
\label{fig:granadosInpaintingResults}
\end{figure}


First of all, we have tested our algorithm on the videos proposed
by Wexler \emph{et al}. \cite{Wexler2007SpaceTime} and Granados \emph{et al}
\cite{Granados2012How}.
The visual results of our algorithm may
be seen in Figures~\ref{fig:wexlerInpaintingResults} and \ref{fig:granadosInpaintingResults}.
We note that the inpainting results of the previous authors on these examples are visually almost perfect, so
very little qualitative improvement can be made. It may be seen that
our results are of similarly high quality to the those of the previous
algorithms. In particular we are able to deal with situations
where several moving objects must be correctly recreated, without
requiring manual segmentation as in \cite{Granados2012How}. We
also achieve these results in at least an order
of magnitude less time than the previous algorithms. 
We note that it is not feasible to apply the method of Wexler \emph{et al}. to
the examples of \cite{Granados2012How}, whose resolution
is too large (up to 1120$\times$754$\times$200 pixels). 

Next, we provide experimental evidence to show the ability of our
algorithm to deal with various situations which appear frequently in real videos,
some of which are not dealt with by previous methods.
Figure~\ref{fig:videoTexture} shows an example of the utility of using
texture features in the inpainting process: without them,
the inpainting result is quite clearly unsatisfactory. We have not directly
compared these results with previous work. However it is quite clear that the
method of \cite{Granados2012How} cannot deal with such situations.
This method supposes that the background is static,
and in the case of dynamic textures, it is not possible to restrict the
search space as proposed in the same method for moving objects. Furthermore, the background
inpainting algorithm of Granados \emph{et al}. \cite{Granados2012Background} supposes that moving background
undergoes a homographic transformation, which is clearly not the case for video textures.
By relying on a plain colour distance between patches, the algorithm of Wexler \emph{et al}.
is likely to produce results similar to the one which may be seen in Figure~\ref{fig:videoTexture}
(middle image).
Finally, to take another algorithm of the literature, the method
of Patwardhan \emph{et al}. \cite{Patwardhan2007Video} would encounter the same problems
as that of \cite{Granados2012Background}, since they copy-and-paste pixels directly
after compensating for a locally estimated motion. More examples of videos
containing dynamic textures can be seen
at \verb|http://www.telecom-paristech.fr/~gousseau/video_inpainting|.

Our algorithm's capacity to deal with moving background is
illustrated by Figure \ref{fig:realignmentComparison}.
We do this in the same
unified framework used for all other examples in this paper, whereas a specific algorithm is needed by Granados \emph{et al}. \cite{Granados2012Background}
to achieve this. Thus, we see that the same core algorithm (iterative
ANN search and reconstruction) can be used in order to deal with a series
of inpainting tasks and situations. Furthermore, we note that no foreground/background
segmentation was needed for our algorithm to produce satisfactory results. Finally,
we note that such situations are not managed using the algorithm of \cite{Wexler2007SpaceTime}.
Again, examples containing moving backgrounds can be viewed at the referenced website.

The generic nature of the proposed approach represents a significant advantage over previous methods,
and allows us to deal with many different situations without having to resort to
manual intervention or the creation of specific algorithms.

\subsection{Execution times}

One of the goals of our work was to accelerate the video inpainting
task, since this was previously the greatest barrier to development in
this area. Therefore, we compare our execution times to those of
\cite{Wexler2007SpaceTime} and \cite{Granados2012How} in
Table~\ref{tab:inpaintingExecutionTimeResults}.

\begin{table}[!ht]
\footnotesize
\centering
\begin{tabular}{|c | r | r | r | r | r|}
\hline
Algorithm & 
\multicolumn{5}{c|}{\centering{ANN execution times for all occluded pixels at full resolution.}} \\
\cline{2-6}
 & Beach Umbrella & Crossing Ladies & Jumping Girl & Duo & Museum\\
 & {\tiny $264\times 68\times 98$} & {\tiny $170\times 80\times 87$} 
 & {\tiny $300\times 100\times 239$} & {\tiny $960\times 704\times 154$} 
 & {\tiny $1120\times 754\times 200$} \\
 \hline
Wexler (kdTrees) & 985 s & 942 s & 7877 s & - & - \\
Ours (3D PatchMatch) & 50 s & 28 s & 155 s & 29 min & 44 min \\
\hline
\hline
Algorithm & \multicolumn{5}{c|}{Total execution time} \\
\hline
Granados & 11 hours & - & - & - & 90.0 hours \\
Ours  & 24 mins & 15 mins & 40 mins & 5.35 hours & 6.2 hours \\
Ours w/o texture & 14 mins &
12 mins & 35 mins & 4.07 hours & 4.0 hours \\
\hline
\end{tabular}
\caption{{\bf Partial and total execution times on different examples}. The partial
inpainting times represent the time taken for the ANN search for all occluded patches
at the full resolution.
Note that for the ``museum''
example, Granados's algorithm is parallelised over the different occluded objects and the background, whereas ours is not.}
\label{tab:inpaintingExecutionTimeResults}
\end{table}

Comparisons with Wexler's algorithm should be obtained carefully,
since several crucial parameters are not specified. In particular,
the ANN search scheme used by Wexler \emph{et al}. requires a parameter,
$\varepsilon$, which determines the accuracy of the ANNs. More formally,
if $W_p$ is a source patch, $W_q$ is the exact NN of
$W_p$ and $W_r$ is an ANN of $W_p$, then the work of \cite{Arya1993Approximate}
guarantees that $d(W_p,W_r) \leq (1+\varepsilon)d(W_p,W_q)$. This parameter has a
large influence on the computational times. For our comparisons,
we set this parameter to 10, which produced ANNs with
a similar average error per patch component as our spatio-temporal
PatchMatch. Another parameter which is left unspecified
by Wexler \emph{et al}. is the number of iterations of ANN search/reconstruction steps
per pyramid level. This has a very large influence on the total execution time.
Therefore, instead of comparing total execution times we simply compare
the ANN search times, as this step represents the majority of the computational load.
We obtain a speedup of 20-50 times over the method of \cite{Arya1993Approximate}.
We also include our total execution times, to give a general idea of the time
taken with respect to the video size. These results show a speedup of
around an order of magnitude with respect to the semi-automatic methods of Granados \emph{et al} \cite{Granados2012How}.
In Table~\ref{tab:inpaintingExecutionTimeResults},
we have also added our execution times without the use of texture features
to illustrate the additional computational load which this adds.

These computation times show that our algorithm is clearly faster than
the approaches of \cite{Wexler2007SpaceTime} and \cite{Granados2012How}. This
advantage is significant because not only is the algorithm more practical
to use, but it is also much easier to experiment and therefore make
progress in the domain of video inpainting.

\section{Further work}
Several points could be improved upon in the present paper.
Firstly, the case where a moving object is occluded for long periods remains very difficult
and is not dealt with in a unified manner here. The one solution to this problem
(temporal subsampling) does not perform well when complex motion is present.
Therefore, other solutions could be interesting to explore.
Secondly, we have observed that using a multi-resolution
texture feature pyramid produces very interesting results. Therefore,
we could perhaps enrich the patch space with other features, such as spatio-temporal
gradients.
Finally, it is acknowledged that
videos of high resolutions still take quite a long time to process (up to several hours).
Further acceleration could be achieved by dimensionality-reducing
transformations of the patch space.

\section{Conclusion}
\label{sec:conclusion}

In this paper, we have proposed a non-local patch-based approach
to video inpainting which produces good quality results
in a wide range of situations, and on high definition videos,
in a completely automatic manner. Our extension of the PatchMatch ANN
search scheme to the spatio-temporal case reduces the time complexity of the
algorithm, so that high definition videos can be processed. We have
also introduced a texture feature pyramid which ensures that dynamic video
textures are correctly inpainted. The case of mobile cameras
and moving background is dealt with by using a global, affine estimation
of the dominant motion in each frame.

The resulting algorithm performs well in a variety of situations,
and does not require any manual input or segmentation. In particular,
the specific problem of inpainting textures in videos has been addressed,
leading to much more realistic results than other inpainting algorithms.
Video inpainting has yet not been extensively used, in a
large part due to prohibitive execution times and/or necessary manual
input. We have directly addressed this problem in the present work.
We hope that this algorithm will make video inpainting
more accessible to a wider community, and help it to
become a more common tool in various other domains, such as video post-production,
restoration and personal video enhancement.

\section{Acknowledgments}

The authors would like to express
their thanks to Miguel Granados for his kind
help, for making his video content publicly available
and for answering several questions concerning his work.
The authors also thank Pablo Arias, Vicent Caselles and Gabriele Facciolo for
numerous insightful discussions and for their help in understanding their
inpainting method.

\appendix
\section{On the link between the non-local patch-based and shift map-based formulations}
\label{sec:variationalGraphCutsLink}

In the inpainting literature, two of the main approaches which
optimise a global objective functional include
non-local patch-based methods \cite{Arias2011Variational,Wexler2007SpaceTime}, which are closely
linked to Non-Local Means denoising \cite{Buades2005Nonlocal}, and shift map-based
formulations such as \cite{Granados2012How,Liu2013ExemplarBased,Pritch2009Shiftmap}.
We will show here that the two formulations are closely linked, and in
particular that the shift map formulation is a specific case of the very
general formulation of Arias \emph{et al}. \cite{Arias2011Variational,Arias2012Analysis}.

As far as possible we keep the notation introduced previously. In particular $p$
is a position in $\mathcal{H}$, $q$ is a position
in $\mathcal{N}_p$ and $r$ is a position in $\tilde{\mathcal{D}}$.
In its most general version, the variational formulation of Arias \emph{et al}. is not based on a one-to-one shift map, but rather on a positive weight function $w:\Omega\times\Omega\rightarrow\mathbb{R}^+$ that is constrained by $\sum_r w(p,r)=1$ to be a probability distribution for each point $p$. Inpainting is cast as the minimisation of the following energy functional (that we rewrite here in its discretised version):
\begin{equation}
E_{\mathrm{arias}}(u,w) = 
\sum_{p \in \cH} \left[\sum_{r \in \tilde{\mathcal{D}}} w(p,r) d^2( W_p , W_{r} )
+ \gamma \sum_{r \in \tilde{\mathcal{D}}} w(p,r)\log{w(p,r)}\right],
\label{eq:energyArias}
\end{equation}
with $\gamma$ a positive parameter.
The first term is a ``soft assignment" version of (\ref{eq:wexlerInpaintingEnergy}), while the second term is a regulariser that favors large entropy weight maps.

Arias  \emph{et al}.  propose the following patch distance:
\begin{equation}
d^2(W_p,W_r) = \sum_{q \in \mathcal{N}_p} g_a(p-q) \varphi[ u(q)-u(r + (p-q))],
\end{equation}
where $g_a$ is the centered Gaussian with standard
deviation $a$ (also called the intra-patch weight function), and $\varphi$
is a squared norm. This is a very general and flexible formulation.

Arias \emph{et al}. optimise this function using an alternate
minimisation over $u$ and $w$, and derive solutions for various
patch distances.
Let us choose the $\ell^2$ distance for $d(W_p , W_{r})$, as is the case
in many inpainting formulations (and in particular the one which
we use in our work).
In this case, the minimisation scheme leads to the following
expressions:
\begin{align}
w(p,r) & = \frac{1}{Z_p}\exp\left( -\frac{d^2(W_p,W_{r})}{\gamma} \right), \label{eq:ariasVariationalSolutions} \\
u(p)   & = \sum\limits_{q \in \mathcal{N}_p} g_a(p-q) \Big(\sum\limits_{r \in \tilde{\mathcal{D}}} w(q,r) u(r+(p-q))\Big).  \label{eq:ariasVariationalSolutionsU}
\end{align}

The parameter $\gamma$ controls the selectivity of the weighting
function $w$. Let us consider the case where each weight function $w(p,.)$ is equal to a single Dirac centered at a single match $p+\phi(p)$.
If, in addition, we consider
that the intra-patch weighting is uniform, in other words $a = \infty$, the cost function
$E_{\mathrm{arias}}$ reduces to:
\begin{equation}
E_{\mathrm{arias}}(u,\phi) = \sum_{p \in \mathcal{H}} \sum_{q \in \mathcal{N}_p} \|u(q) - u(q+\phi(p))\|_2^2,
\label{eq:ariasReducedWexler}
\end{equation}
which is the formulation of Wexler \emph{et al}. \cite{Wexler2007SpaceTime}.

Rewriting \eqref{eq:ariasVariationalSolutionsU} in the particular case just described, yields the optimal inpainted image
\begin{equation}
 u(p) = \frac{1}{|\mathcal{N}_p|}\sum_{q\in\mathcal{N}_p} u(p+\phi(q)), \; \forall p \in \mathcal{H},
\label{eq:ariasReconstruction}
\end{equation}
as the (aggregated) average of the examples indicated by the NNs of the patches
which contain $p$. Suppose that each pixel is reconstructed using the NN of the patch centred on it:
\begin{equation}
u(p) = u(p+\phi(p)), \; \forall p \in \mathcal{H},
\label{eq:shiftMapReconstruction}
\end{equation}
as is the case in the shift map-based formulations.
Then the functional becomes\footnote{We note that
using the reconstruction of Equation~\ref{eq:shiftMapReconstruction}
poses problems on the occlusion border, but we ignore this here for the
sake of simplicity and clarity.}:
\begin{equation}
E_{\mathrm{arias}}(u,\phi) = \sum_{p \in \mathcal{H}} \sum_{q \in \mathcal{N}_p} \| u(q + \phi(q)) - u(q + \phi(p)) \|_2^2,
\end{equation}
\noindent which effectively depends only upon $\phi$.

If we look at the shift map formulation proposed by
Pritch \emph{et al}., we find the following cost function over the shift map only:
\begin{equation}
E_{\mathrm{pritch}}(\phi) = 
\sum_{p \in \mathcal{H}} \sum_{q \in \mathcal{N}_p}\Big(\|u(q) - u(q + \phi(p))\|_2^2
+ \|\nabla u(q) - \nabla u( q + \phi(p))\|_2^2\Big).
\label{eq:shiftMapSmoothEnergy}
\end{equation}
Let us consider the first part, concerning the image colour values.
Since we have $u(p) = u(p+\phi(p))$, we obtain again:
\begin{equation}
\label{eq:shiftMapSmoothEnergy2}
E_{\mathrm{pritch}}(\phi) = \sum_{p \in \mathcal{H}} \sum_{q \in \mathcal{N}(p)}\|u(q + \phi(q)) - u(q + \phi(p))\|_2^2.
\end{equation}
Thus, the shift map cost function may be seen as a special
case of the non-local patch-based formulation of Arias \emph{et al}. under the following conditions :

\begin{itemize}
\item $d^2(W_p,W_{r}) = \sum_{q \in \mathcal{N}_p} || u(q) - u(r+ (p-q)) ||^2_2$;
\item $\gamma=0$;
\item The intra-patch weighting function $g_a$ is uniform;
\item $u(p) = u(p+\phi(p))$, that is, $u(p)$ is reconstructed using its correspondent according to a single shift map.
\end{itemize}

Arias \emph{et al}. \cite{Arias2012Analysis} have shown
the existence of optimal correspondance maps for a relaxed version of Wexler's energy
(Equation~\ref{eq:energyArias}), and also that the minima of
Equation~\ref{eq:energyArias}) converge to the minima of Equation~\ref{eq:ariasReducedWexler}
as $\gamma \rightarrow 0$. Such results also highlight the link between
different inpainting energies.

However, one should keep in mind that the two formulations which we
have considered (that of Arias \emph{et al}. and Pritch \emph{et al}.)
are certainly not equivalent, for reasons
such as the difference in optimisation methodology and the
presence of a gradient term in the formulation of Pritch \emph{et al}.
Furthermore, the choice of the reconstruction $u(p) = u(p+\phi(p))$ was 
not considered by Arias \emph{et al}., meaning that results may differ.


\section{Comparing textured patches}
\label{sec:texturedPatchesAnalysis}

In this Appendix, we look in further detail at
the reasons why textures may pose a problem when
comparing patches for the purposes of inpainting.
Liu and Caselles noted in \cite{Liu2013ExemplarBased} that the
subsampling necessary for the use of multi-resolution pyramids
inevitably entails a loss of detail, leading to difficulties
in correctly identifying textures. In fact,
we found that this difficulty may occur
at \emph{all} the pyramid levels in images and videos.
Roughly speaking, we observed that textured patches are
quite likely to be matched with smooth ones. The following
simple computations quantify this phenomenon.

\begin{table}
\begin{tabular}{llllllll}
\hline
Patch size & $3\times 3$ & $5\times 5$ & $3\times 3\times 3$ & $7\times 7$ & $9\times 9$ & $11\times 11$ & $5\times 5\times 5$  \\
\hline
Probability & $8\times 10^{-2}$ & $10^{-2}$ & $6\times 10^{-3}$ & $4.1\times 10^{-4}$ & $5.5\times 10^{-6}$ & $3\times 10^{-7}$ & $2\times 10^{-7}$ \\
\hline
\end{tabular}
\caption{{\bf Probability of producing a random 2D or 3D patch that is closer to a random
reference patch than to a constant one with same mean value}. Values are obtained through 
numerical simulations averaged over ten runs for each experiment. Components of random patches
are i.i.d. according to the centred normal law with a grey level variance of 25.}
\label{tab:randomPatchNumericalSimulationImageCase}
\end{table}

\subsection{Comparing patches with the classical $\ell^2$ distance}

The first reason concerns the patch distance.
Let us consider a white noise patch, $W$, which is a vector
of i.i.d. random variables $W_1 \cdots W_N$, where $N$ is the number of
components in the patch (number of pixels for grey-level patches), and the distribution of all $W_i$'s is $f_W$.
Let $\mu$ and $\sigma^2$ be, respectively, the average and variance
of $f_W$. Let us consider another random patch $V$ following same distribution, and the constant patch $Z$, composed of $Z_i = \mu$, $i=1 \cdots N$.

In this simple situation, we see that $\mathbb{E}[\|W-V\|^2_2] = 2\mathbb{E}[\|W-Z\|^2_2]$.
Therefore, on average, the sum-of-squared-differences (SSD) between two patches of the same distribution
is \emph{twice} as great as the SSD between a randomly distributed patch and a
constant patch of value $\mu$.

The previous remark is only valid on average between
three patches $W$, $V$ and $Z$. In reality,
we have many random patches $V$ to choose from, and
it is sufficient that one of these be better than $Z$ for the
least patch distance to identify a ``textured'' patch.
Therefore, a more interesting question is the following. Given a white noise
patch $W$, what is the probability that the patch $V$ will
be better than the constant patch $Z$.
This is slightly more involved, and we shall limit ourselves to the
case where $W$ and $V$ consist of i.i.d. pixels with a normal distribution.

The SSD patch distance between $W$ and $V$ follows a chi-square distribution $\chi^2(0,2 \sigma ^2)$,
and that between $W$ and $Z$ follows $\chi^2(0,\sigma ^2)$.
With this, we may numerically compute the probability of a random patch being better
than a constant one. Since the chi-squared law
is tabulated, it is much the same thing to use numerical simulations.

In Table~\ref{tab:randomPatchNumericalSimulationImageCase},
we show the corresponding numerical values for both 2D (image) and 3D (video) patches.
It may be seen that for a patch of size $9 \times 9$, there is very little
chance of finding a better patch than the constant patch.
In the video case, we see that in the case of $5 \times 5 \times 5$ patches,
there is a $2\times 10^{-7}$ probability of creating a better patch randomly.
This corresponds to needing an area of $170 \times 170 \times 170$ pixels in a video
in order to produce on average one better random patch.
While this is possible, especially in higher-definition videos,
it remains unlikely for many situations.

The question naturally arises of why the problem of comparing textures has not
been more discussed in the patch-based inpainting literature. Indeed, to the best of our
knowledge, only Bugeau \emph{et al}. \cite{Bugeau2010Comprehensive} and Liu
and Caselles \cite{Liu2013ExemplarBased} have clearly identified this problem
in the case of image inpainting.
This is due to the fact that most other inpainting algorithms
\emph{restrict the ANN search space} to a local neighbourhood around the
occlusion. 
Unfortunately, this restriction principle does not hold in video inpainting
since the information can be found anywhere in the video volume, in
particular when a complex movement must be reconstructed.


\begin{figure}
\begin{tabularx}{\linewidth}{@{} *4{>{\centering\arraybackslash}X}@{}}
\includegraphics[width = \linewidth]{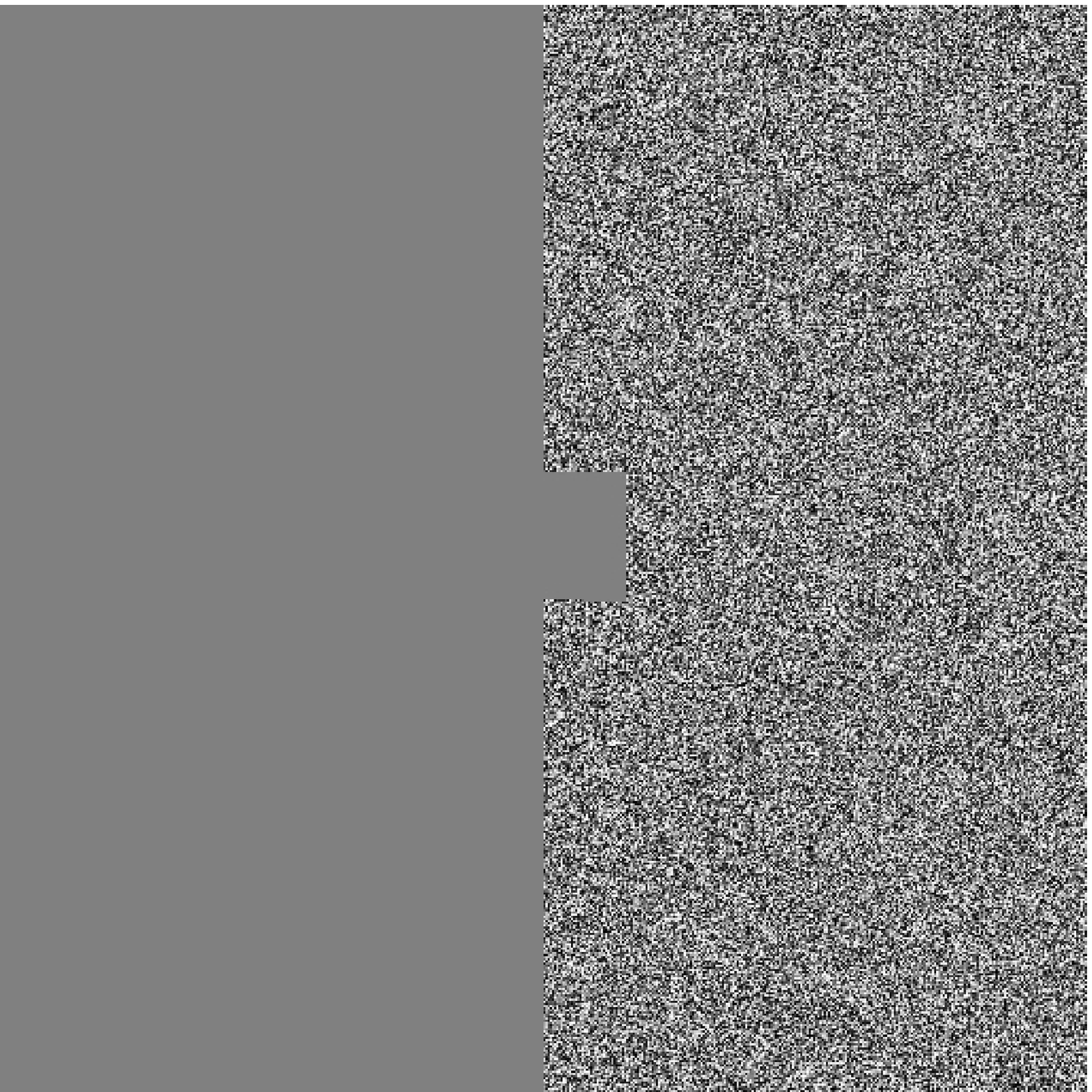} &
\includegraphics[width = \linewidth]{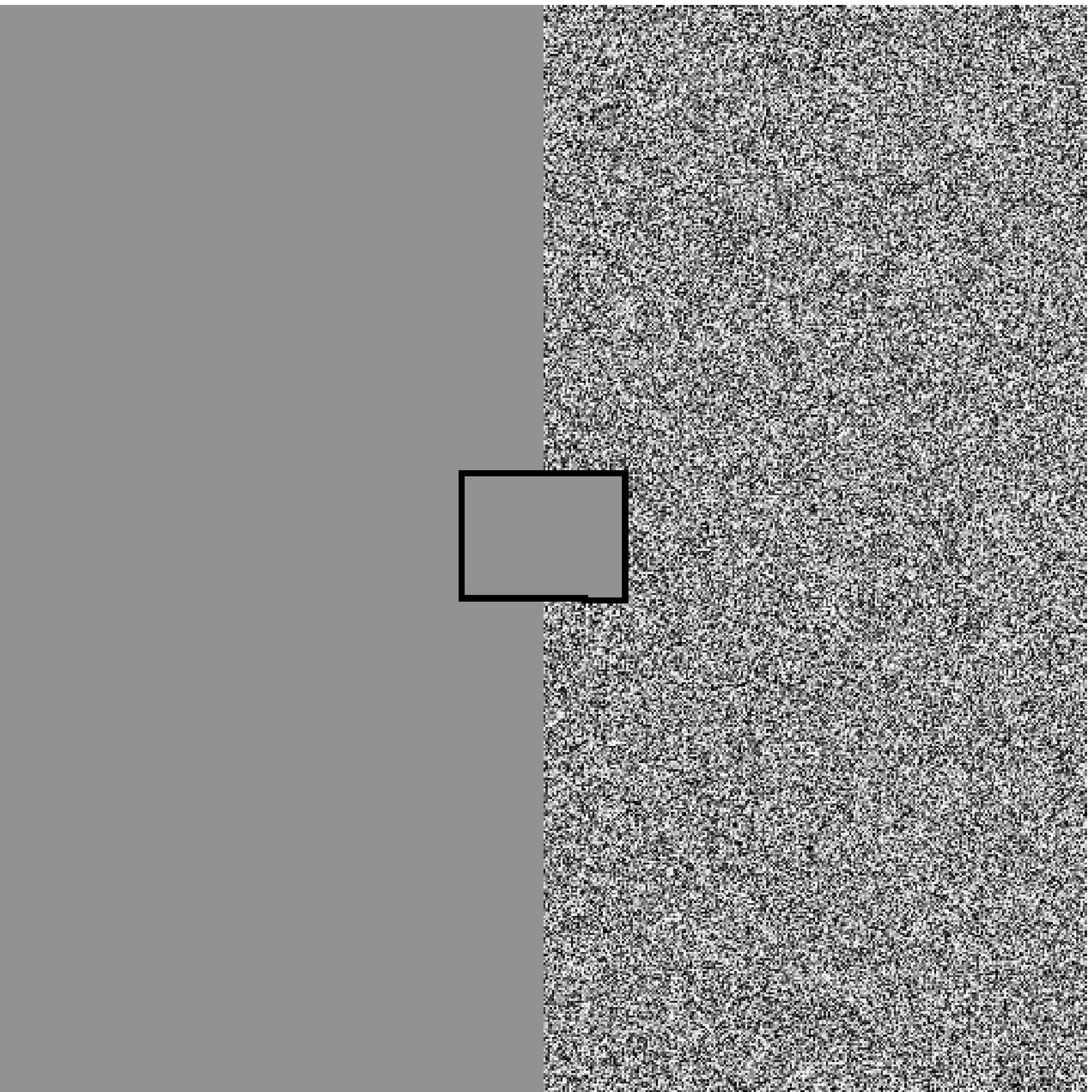} &
\includegraphics[width = \linewidth]{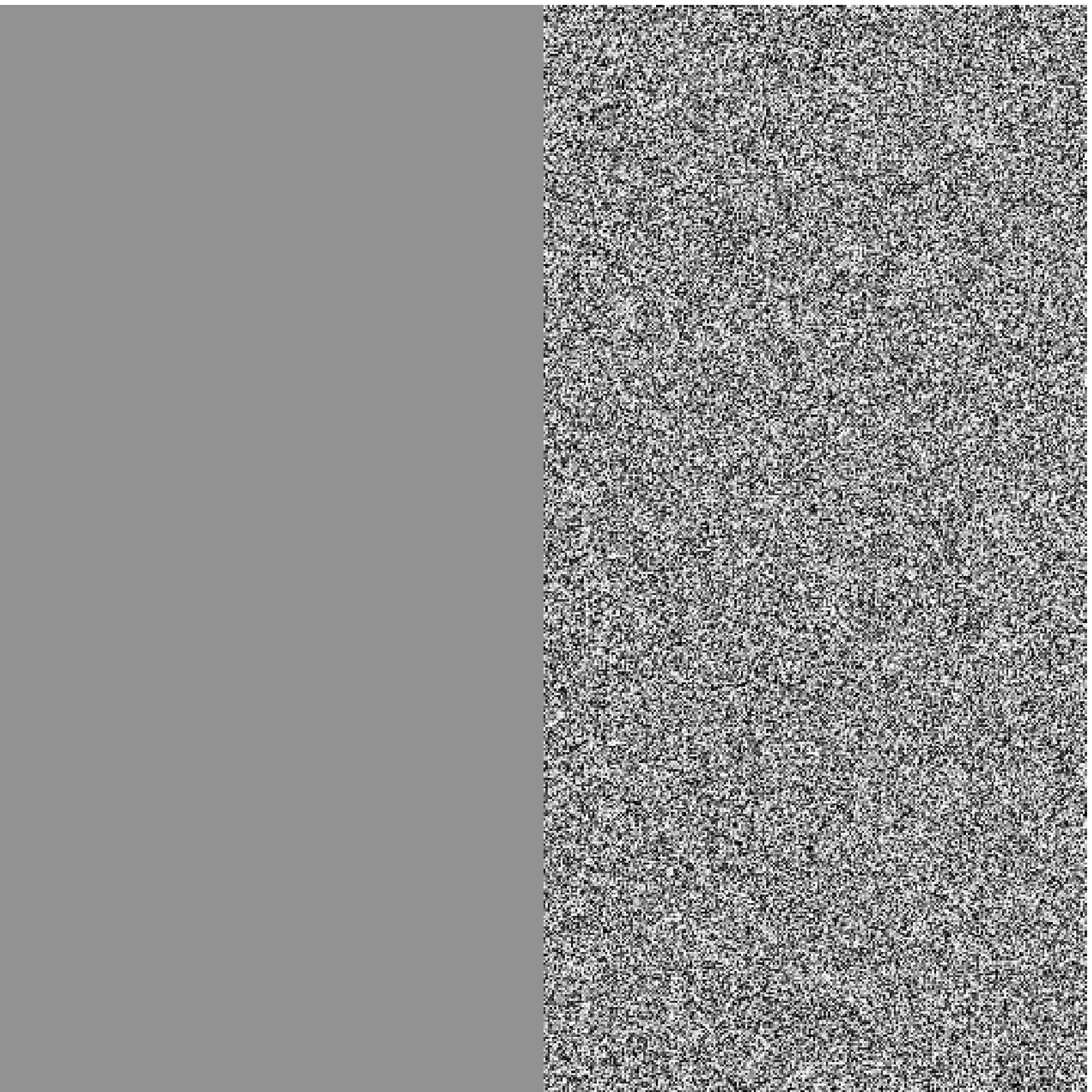} &
\includegraphics[width = \linewidth]{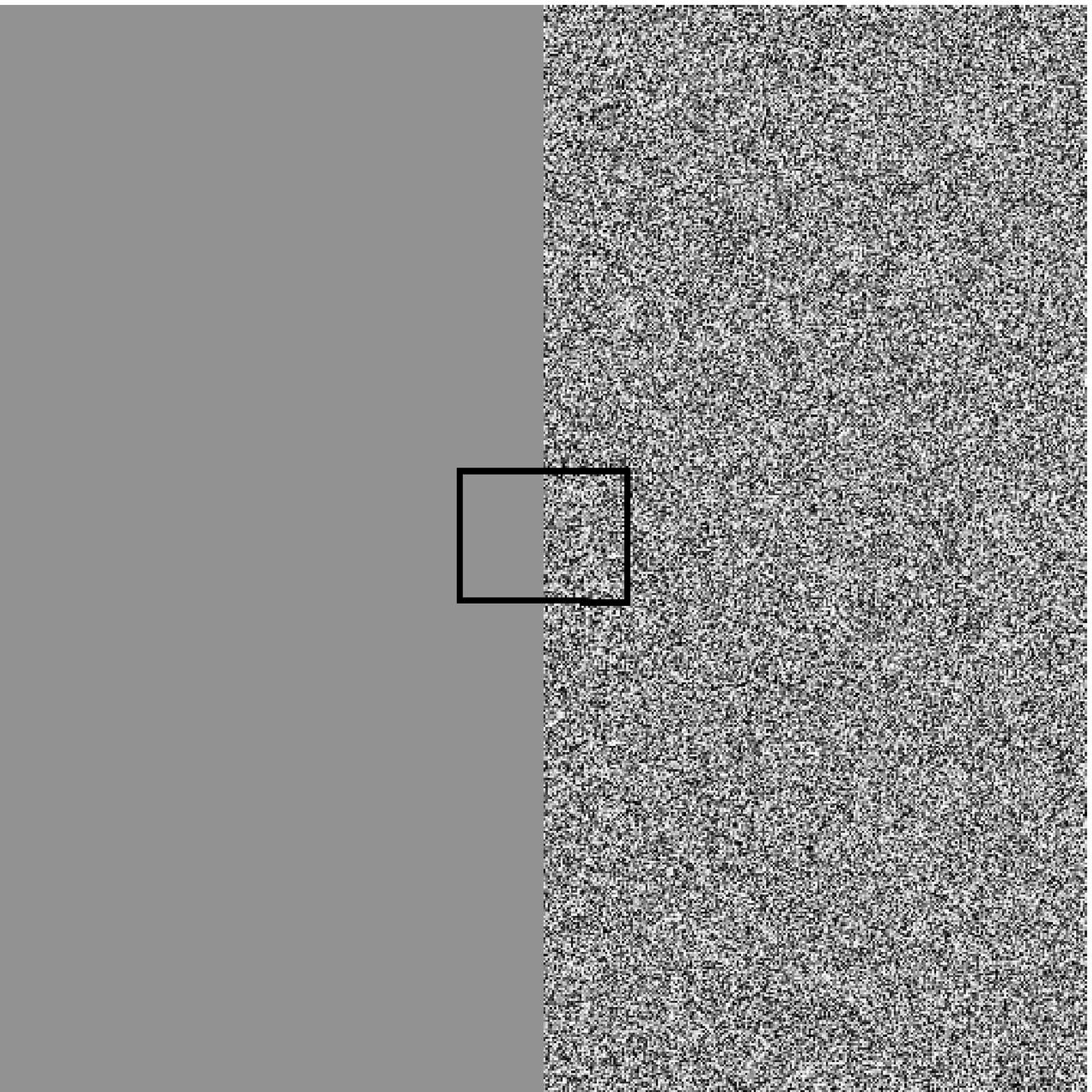} \\
Inpainting without texture features & Inpainting without texture features, with occlusion border outlined &
Inpainting with texture features & Inpainting with texture features, with occlusion border outlined \\
\end{tabularx}
\caption{{\bf A toy example of the utility of the textures features}. With them, we are able to distinguish between white
noise (right) and the constant area (left), and thus recreate the noise.}
\label{fig:inpaintingNoise}
\end{figure}


\subsection{ANN search with PatchMatch}

We have indicated that the $\ell^2$ grey-level/colour patch distance is problematic for inpainting
in the presence of textures.
Additionally, this problem is exacerbated by the use of PatchMatch. Indeed,
the values of $\phi$ which lead to textures are not piecewise
constant, and are therefore not propagated through $\phi$ during
the PatchMatch algorithm. On the other hand, smooth patches represent
on average a good compromise as ANNs and are the shifts which
lead to them are piecewise constant and therefore well propagated throughout
$\phi$.
Another problem which may lead to smooth patches being used is the weighted average
reconstruction scheme. This can lead to blurry results which in turn means that
smooth patches are identified.

One solution to these problems is the use of our texture feature pyramid (\S \ref{subsec:inpaintingTextures}).
This pyramid is inpainted simultaneously with the colour video pyramid, and
thus helps to guide the algorithm in the choice of which patches to use
for inpainting.

Figure~\ref{fig:inpaintingNoise} shows an interesting situation:
we wish to inpaint a region which contains white noise. This toy example serves
as an illustration of the appeal of our texture features. Indeed, it is quite
clear that without them, there is no chance of inpainting the occlusion
in a manner which would seem ``natural'' to a human observer, whereas
with them it is possible, in effect, to ``inpaint noise''.

\bibliographystyle{siam}
\bibliography{inpaint}

\end{document}